\documentclass{article}

\usepackage{microtype}
\usepackage{graphicx}
\usepackage{subcaption}
\usepackage{booktabs} % for professional tables

\usepackage{hyperref}

\usepackage[preprint]{icml2026}

\usepackage{amsmath}
\usepackage{amssymb}
\usepackage{mathtools}
\usepackage{amsthm}

\usepackage[capitalize,noabbrev]{cleveref}

\theoremstyle{plain}

\theoremstyle{definition}

\theoremstyle{remark}

\usepackage[textsize=tiny]{todonotes}
\usepackage[most]{tcolorbox} % Include tcolorbox package
\usepackage{xcolor}

\usepackage{array}    % 用于自定义列格式
\usepackage{makecell} % 用于换行单元格内容
\usepackage{multirow} % 用于跨行合并
\usepackage{adjustbox} % 可选：自动缩放表格

\tcbuselibrary{breakable}

\icmltitlerunning{TwiFF (Think With Future Frames)}

\begin{document}

\twocolumn[
  \icmltitle{TwiFF (Think With Future Frames): \\
  A Large-Scale Dataset for Dynamic Visual Reasoning}



  \icmlsetsymbol{equal}{*}

  \begin{icmlauthorlist}
    \icmlauthor{Junhua Liu}{zju}
    \icmlauthor{Zhangcheng Wang}{4pd}
    \icmlauthor{Zhike Han}{hzcu,zju}
    \icmlauthor{Ningli Wang}{AIMS}
    \icmlauthor{Guotao Liang}{buaa}
    \icmlauthor{Kun Kuang}{zju}
  \end{icmlauthorlist}

  \icmlaffiliation{zju}{College of Computer Science and Technology, Zhejiang University, Hangzhou, China}
  \icmlaffiliation{4pd}{4paradigm, Beijing, China}
  \icmlaffiliation{hzcu}{School of Computer and Computing Science, Hangzhou City University, Hangzhou, China}
  \icmlaffiliation{AIMS}{Henan Academy of Innovations in Medical Science (AIMS), Zhengzhou, China}
  \icmlaffiliation{buaa}{School of Software, Beihang University, Beijing, China}

  \icmlcorrespondingauthor{Zhike Han}{hanzk@hzcu.edu.cn}
  \icmlcorrespondingauthor{Ningli Wang}{wningli@vip.163.com}

  \icmlkeywords{Visual Chain-of-Thought, Interleave Text and Image Generation, Multimodal Reasoning}

  \vskip 0.3in
]



\printAffiliationsAndNotice{}  % no special notice (required even if empty)
\definecolor{softgreen}{HTML}{43A047}
\definecolor{softred}{HTML}{D32F2F}

\begin{abstract}
Visual Chain-of-Thought (VCoT) has emerged as a promising paradigm for enhancing multimodal reasoning by integrating visual perception into intermediate reasoning steps. However, existing VCoT approaches are largely confined to static scenarios and struggle to capture the temporal dynamics essential for tasks such as instruction, prediction, and camera motion. To bridge this gap, we propose TwiFF-2.7M, the first large-scale, temporally grounded VCoT dataset derived from $2.7$ million video clips, explicitly designed for dynamic visual question and answer. Accompanying this, we introduce TwiFF-Bench, a high-quality evaluation benchmark of $1,078$ samples that assesses both the plausibility of reasoning trajectories and the correctness of final answers in open-ended dynamic settings. Building on these foundations, we propose the TwiFF model, a unified modal that synergistically leverages pre-trained video generation and image comprehension capabilities to produce temporally coherent visual reasoning cues-iteratively generating future action frames and textual reasoning. Extensive experiments demonstrate that TwiFF significantly outperforms existing VCoT methods and Textual Chain-of-Thought baselines on dynamic reasoning tasks, which fully validates the effectiveness for visual question answering in dynamic scenarios. Our code and data is available at \url{https://github.com/LiuJunhua02/TwiFF}.
\end{abstract}
\vskip 0.3in
\definecolor{correctblue}{HTML}{46B1E1}
\definecolor{factualred}{HTML}{FF0000}

\section{Introduction}
\label{sec:intro}
\begin{figure}[t]
  \centering
  \centerline{\includegraphics[width=\linewidth]{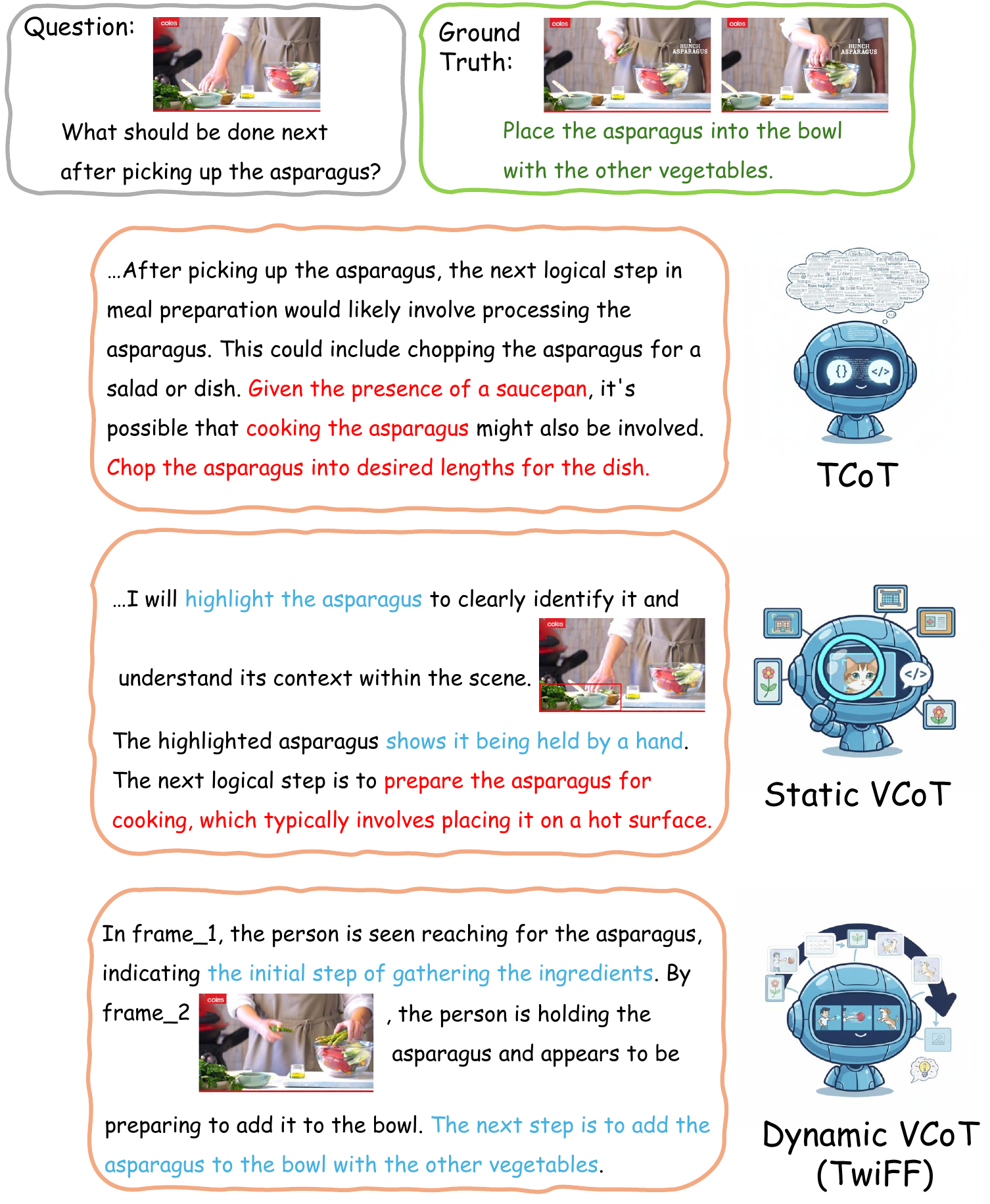}}
  \caption{Dynamic VCoT (TwiFF) vs. TCoT and Static VCoT: \textcolor{correctblue}{Blue} denotes correct model outputs, and \textcolor{factualred}{red} denotes errors or factual inconsistencies.}
  \label{fig:motivation}
\end{figure}
Since Textual Chain-of-Thought (TCoT) relies solely on textual reasoning, it performs poorly on tasks that require deep visual iterative processing~\cite{su2025thinking}. Researchers have attempted to construct a Visual Chain-of-Thought (VCoT), enabling multimodal models to actively perceive, process, and utilize visual information as intermediate steps in their cognitive reasoning process, thereby enhancing their capability to tackle complex visual reasoning tasks. Current state-of-the-art multimodal large language models~\cite{hu2024visual,zheng2025deepeyes,fu2025refocus,zhang2025thyme} implement VCoT by invoking external tools to perform diverse operations on the input image. Although recent studies~\cite{li2025zebra,li2025imagine,gu2025thinkmorph,cheng2026omni,zhang2025cooper,shi2025mathcanvas} have explored generating visual reasoning cues directly within a unified model by leveraging its intrinsic image generation capabilities, their VCoT training data is still constructed via tool-augmented pipelines. Consequently, these models remain fundamentally constrained to reasoning about only the visual content present in the input image, rendering them primarily effective for visual question answering in static scenarios, such as maze navigation, geometric reasoning, visual search, and jigsaw assembly. In contrast, the VCoT developed for such settings performs poorly in dynamic scenarios, including instructional (guiding user actions step-by-step), predictive (forecasting future events or outcomes), or camera (prescribing desired camera motions) task, as these tasks require reasoning about future events or anticipating visual states that extend beyond the current observation. 

In dynamic scenarios, effective VCoT requires reasoning about future or intermediate visual states, rather than relying solely on the contextual information present in the input image. Previous studies have found that generating video-based predictions of physical-world actions can effectively enhance model accuracy in scenarios such as robotic action decision-making~\cite{zhao2025cot, yang2024position} and navigation~\cite{dong2025unified}. Building on this insight, we argue that the key to enabling VCoT with a unified model in open-ended dynamic scenarios lies in jointly leveraging its pre-trained video generation and image comprehension capabilities: reasoning should first trigger the generation of images depicting critical action steps, which are then enriched with textual descriptions to articulate the underlying reasoning process, ultimately leading to more accurate predictions.

To enable unified models to perform temporally coherent visual reasoning, we introduce TwiFF-2.7M, a visual question-answering dataset constructed from $2,708,318$ distinct video clips, featuring interleaved visual-textual reasoning trajectories grounded in dynamic scenarios. It spans three major domains-instructional, predictive, and camera-each encompassing a diverse range of topic and task. To the best of our knowledge, TwiFF-2.7M is the first VCoT dataset explicitly structured along the temporal dimension across such a broad spectrum of dynamic scenarios. By deriving reasoning paths directly from authentic video segments, TwiFF-2.7M ensures strong alignment between its reasoning steps and the physical laws governing the real world. As illustrated in the~\cref{fig:motivation}, the dynamic VCoT generates responses that better align with the plausible progression of events in real-world videos, whereas TCoT and Static VCoT are prone to hallucinations or produce erroneous and implausible reasoning steps during inference.

We also introduce TwiFF-Bench, a high-quality benchmark comprising $1,078$ carefully curated samples specifically designed to comprehensively evaluate both the plausibility of a model's reasoning process and the correctness of its final answers in dynamic scenarios.

The TwiFF model, trained on TwiFF-2.7M, achieves superior performance on dynamic scenarios reasoning benchmarks compared to both the static VCoT model and the conventional TCoT model. Moreover, our experiments reveal three key findings: (1) The effectiveness of VCoT stems from the synergistic interaction between its textual and visual modalities—neither modality alone suffices to achieve its full reasoning capability. (2) Visual cues that align with real-world physical dynamics are crucial for the model's answer accuracy; conversely, misleading visual cues can significantly degrade the reliability of the model's responses. (3) Visual cues in dynamic scenarios possess inherent information-compression potential, enabling the model to suppress noise irrelevant to the underlying actions.

Our main contributions are as follows:
\begin{itemize}
\item We present \textbf{TwiFF-2.7M}, the first large-scale, dynamic VCoT dataset comprising step-by-step visual-textual rationales derived from over 2.7 million video clips. It is explicitly designed to support dynamic visual reasoning—enabling models to anticipate future states beyond static observations.

\item We introduce \textbf{TwiFF-Bench}, a high-quality evaluation benchmark with $1,078$ samples that jointly evaluates the plausibility of intermediate reasoning steps and the correctness of final predictions in open-ended, temporally evolving scenarios—addressing a critical limitation of existing visual question-answering benchmarks that focus solely on end-task accuracy.

\item We show that \textbf{TwiFF} trained on TwiFF-2.7M outperform both TCoT and static VCoT approaches in dynamic scenario visual question-answering tasks. Our experiments further reveal that physically plausible visual cues not only improve answer accuracy but also effectively preserve task-critical visual information.
\end{itemize}
\begin{figure*}[t!]
  \centering
  \centerline{\includegraphics[width=\linewidth]{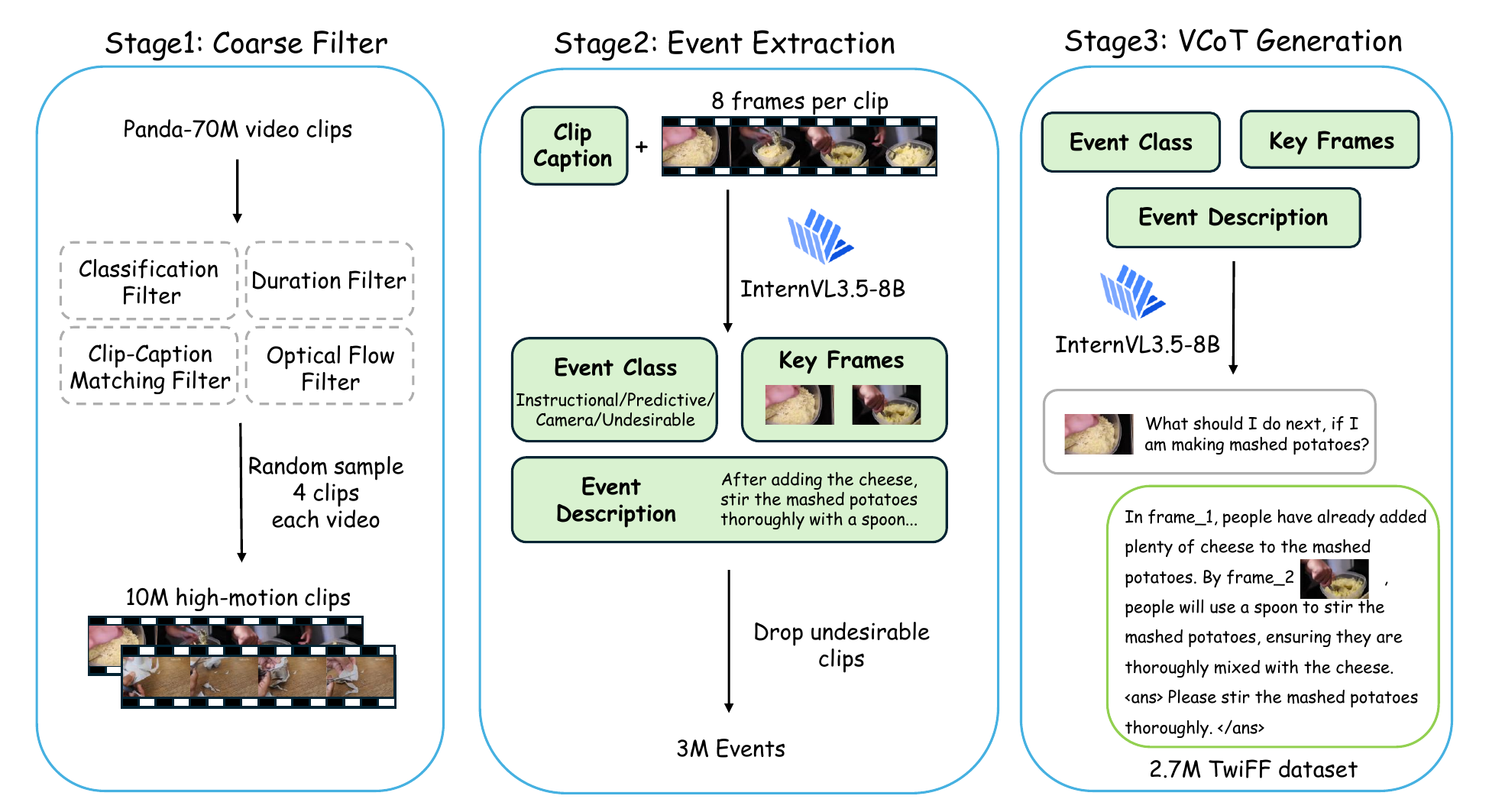}}
  \caption{\textbf{Data generation.} We generate TwiFF-2.7M in three stages.}
  \label{fig:pipeline}
\end{figure*}
\section{Related Work}
\label{sec:relatedwork}
\subsection{Multimodal Reasoning}
Existing approaches to multimodal reasoning primarily fall into two paradigms: text-based TCoT and visual-textual interleaved VCoT. Recent multimodal models—such as InternVL3.5~\cite{wang2025internvl3} and Qwen3VL~\cite{bai2025qwen3}—leverage textual reasoning traces to improve performance on visual question-answering. To incorporate visual cues into the reasoning process, several works augment reasoning with external tools that actively process the input image. DeepEyes~\cite{zheng2025deepeyes}, Refocus~\cite{fu2025refocus}, and Thyme~\cite{zhang2025thyme} implement VCoT by invoking predefined image processing routines during inference. Going a step further, SKETCHPAD~\cite{hu2024visual} leverages additional vision models, such as segmentation and object detection, to generate diverse and richer visual cues, thereby enhancing the reasoning process.

An alternative approach to incorporating visual cues into the reasoning process involves having a unified model directly generate images as part of VCoT. By leveraging tools to synthesize visual reasoning cues and constructing VCoT datasets for training~\cite{shi2025mathcanvas,li2025imagine,zhang2025cooper,gu2025thinkmorph}, these models are endow unified models with the ability to reason visually. 

The aforementioned VCoT reasoning is confined to the content of the input image and does not extend beyond it, rendering it ineffective in dynamic scenarios that require forecasting future events. Although Zebra-CoT~\cite{li2025zebra} incorporates a limited amount of embodied data, the scenarios it covers are relatively narrow and do not adequately explore the role of such data in dynamic reasoning tasks. Therefore, our proposed TwiFF-2.7M effectively addresses the gap in existing research on dynamic reasoning.

\subsection{Multimodal Benchmark}
Most existing multimodal benchmarks~\cite{tong2024eyes, yue2024mmmu, lu2023mathvista, yu2023mm} are primarily confined to evaluating a model's ability to understand image content, largely overlooking its capacity for predicting or planning future events. Although MMBench~\cite{liu2024mmbench} includes a small number of prediction tasks, its limited sample size renders it inadequate as an effective benchmark for evaluating such tasks. While Seed-Bench-R1~\cite{chen2025exploring} is specifically designed for future action prediction, its multiple-choice format reduces both the task's difficulty and realism, as real-world scenarios typically require open-ended responses rather than selection from predefined options. Moreover, Seed-Bench-R1 evaluates models solely based on answer accuracy, overlooking the plausibility and coherence of the reasoning processes underlying their predictions.

\section{Methodology}
\subsection{Data Generation}
%------------------------
To build TwiFF-2.7M, we leverage video clips from Panda-70M~\cite{chen2024panda}, a dataset sourced from YouTube that spans a diverse range of domains. However, not all video clips contain events exhibiting clear causal dependencies, and event-relevant frames are unevenly distributed across clips. To address these challenges, we propose a three-stage pipeline to filter and construct VCoT data, as illustrated in~\cref{fig:pipeline}.

\textbf{Stage1 Coarse Filter.} During this stage, we primarily filtered out low-quality video clips that exhibited insufficient visual variation. Four distinct criteria were applied to perform this filtering process.
First, to ensure the semantic alignment between each clip and its associated caption, we leveraged the Unmasked Teacher~\cite{li2023unmasked} matching score provided by Panda-70M and discarded all clips with a matching score below $0.43$. This threshold guarantees that the captions accurately reflect the visual content of the corresponding clips.
Second, we retained only those clips from Panda-70M labeled as desirable, thereby excluding clips characterized by poor visual quality which are generally unsuitable for constructing effective VCoT data, such as static foreground images, screen-in-screen compositions, or computer screen recordings.
Third, to ensure that each clip captures at least one complete event or action, we removed all clips with a duration shorter than 2 seconds.
Fourth, to ensure perceptible motion or visual change within each clip, we filtered out videos with insufficient dynamics. Specifically, for each candidate clip, we uniformly sampled 8 frames and computed the mean optical flow magnitude between each pair of consecutive frames (see~\cref{APP:OpticalFlowComp} for details). Clips whose maximum inter-frame optical flow magnitude was below 4 were excluded as lacking dynamic content.
Finally, to promote source diversity and prevent overrepresentation of any single video, we retained at most four qualifying clips per original video. This multi-stage filtering pipeline yielded a final dataset of $10,596,462$ high-dynamics video clips.

\textbf{Stage2 Event extraction.} In this stage, we leverage a multimodal large language model (e.g. InternVL3.5-8B \cite{wang2025internvl3}) to process highly dynamic video clips and extract salient events along with key frames. Each clip is classified into one of four categories: \textit{Instructional} (procedural demonstrations such as cooking or mechanical assembly), \textit{Predictive} (events requiring causal or temporal reasoning), \textit{Camera} (highlighting cinematographic techniques like tracking shots or zoom-ins), or \textit{Undesirable} (clips lacking clear structure or not fitting the above types), with the latter being discarded. For the remaining clips, the model selects at least two key frames capturing the cause, process, and outcome of the event, and generates detailed textual descriptions. To balance efficiency and diversity, we uniformly sample eight frames per clip for keyframe selection and description generation. After filtering, we retain $3,075,048$ high-quality event instances, each annotated with process-relevant key frames and natural-language narratives.

\textbf{Stage3 VCoT generation.} In the final stage, we use the event data extracted in prior stages to construct dynamic VCoT data. Specifically, for each event, its key frames are temporally ordered according to their occurrence within the event sequence. The earliest frame (denoted as \textit{frame}$_1$) serves as the query image, while subsequent frames constitute the visual cues supporting the reasoning process in the answer. A multimodal large language model is employed to first generate a question conditioned on the visual content of \textit{frame}$_1$ along with the event's category and textual description. Subsequently, the model constructs a reasoning chain in which visual inputs and textual reasoning steps alternate to simulate human-like visual inference, following a structured format:
$\langle \text{reasoning about } \textit{frame}_{i-1} \rangle$,
$\langle \textit{frame}_i \rangle$,
$\langle \text{reasoning about } \textit{frame}_i \rangle$,
$\dots$,
$\langle \text{final answer} \rangle$.

This pipeline ultimately yields a TwiFF-2.7M comprising $2,708,318$ VCoT data. The data generation prompts used in stage2 and stage3 are shown in \cref{APP:dataprompt}.

%-------------------------------------------------------------------------
\subsection{Benchmark generation and evaluation} 
Following the official Panda-70M split, TwiFF-Bench consists of $1,078$ question-answer pairs derived from its test subset using our training data construction pipeline, guaranteeing zero overlap with the training data. To ensure the validity and reliability of the benchmark, all generated samples were manually filtered. Samples exhibiting flawed reasoning, incorrect answers, or overly open-ended responses were systematically filtered out. Following prior work~\cite{jiang2025mme, yao2025argus}, our benchmark employs GPT-5.1 as a judge to evaluate both the reasoning process and the final answer in terms of reasonableness and accuracy. Specifically, we provide the judge model with two reference inputs: a reference VCoT and the ground-truth answer, both of which are derived from actual future events observed in the video—ensuring they are factually grounded and temporally consistent with the visual context. The reasoning trace and final answer generated by the evaluated model are then submitted to the judge for separate scoring of the CoT reasonableness and answer correctness. To promote fairness and reduce bias, we explicitly instruct the judge not to penalize a CoT solely for omitting explicit image references. Instead, the evaluation should prioritize the logical coherence, plausibility, and factual alignment of the reasoning process with respect to the ground-truth future events. All scores are assigned on a scale from 0 to 5.

To investigate our model's generalization to out-of-distribution (OOD) datasets, we evaluate it using Seed-Bench-R1~\cite{chen2025exploring}, a benchmark consisting of $4,676$ samples drawn from EPIC-Kitchens-100~\cite{damen2022rescaling} and Ego4D~\cite{grauman2022ego4d}. To maintain consistency with the TwiFF-Bench and better reflect real-world scenarios, we formulate open-ended questions by providing the model only with the current observation from Seed-Bench-R1 together with the question itself. Unlike TwiFF-Bench, Seed-Bench-R1 does not include reference reasoning traces. Therefore, the  judge model focuses exclusively on the accuracy of the final answers.

\begin{figure*}[t!]
  \centering
  \centerline{\includegraphics[width=\linewidth]{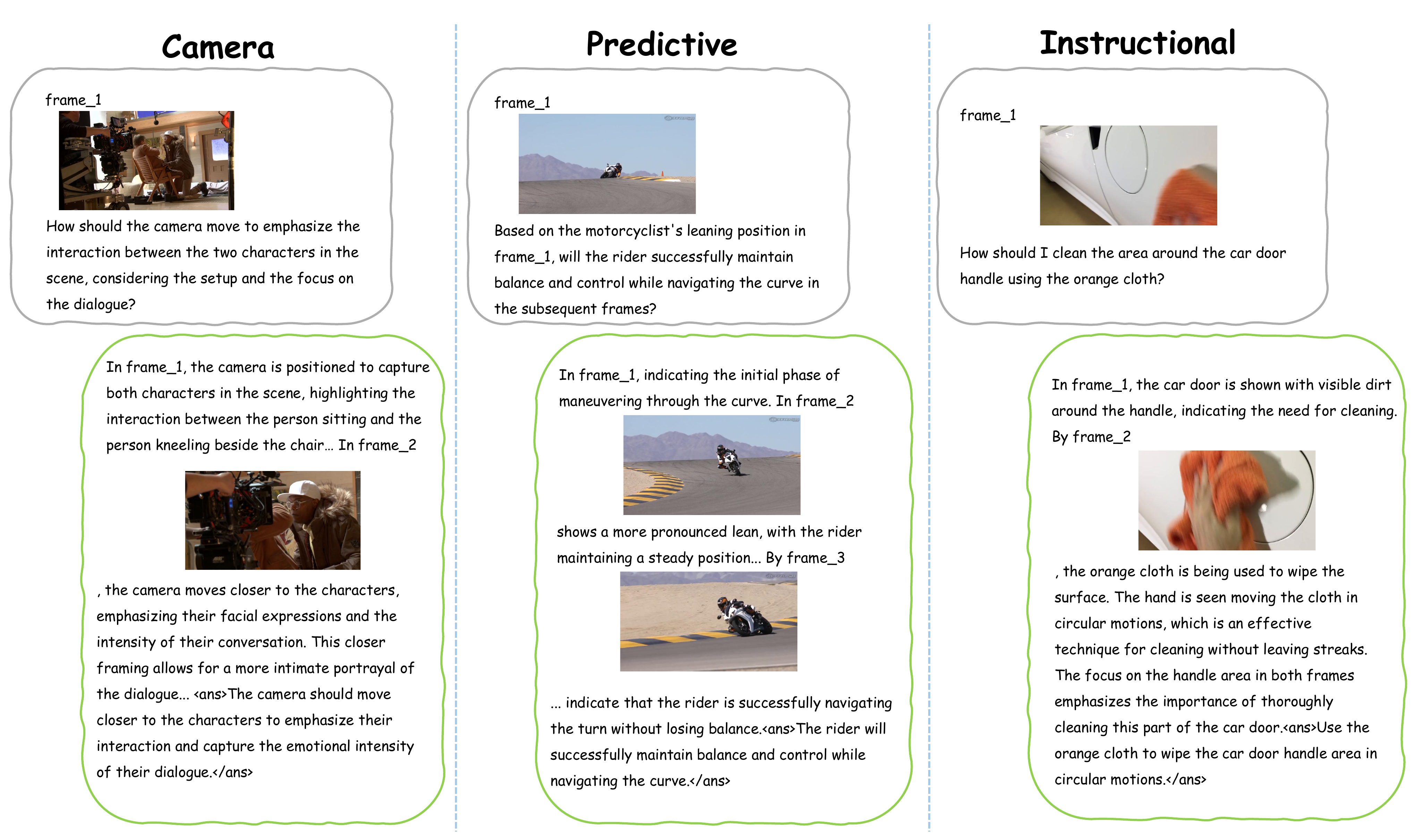}}
  \caption{\textbf{TwiFF Overview.} Both TwiFF-2.7M and TwiFF-Bench consist of dynamic VCoT with interleaved image-text reasoning, covering three task types: instructional, predictive, and camera.}
  \label{fig:datasetoverview}
\end{figure*}
\begin{figure}[t!]
  \centering
  \centerline{\includegraphics[width=\columnwidth]{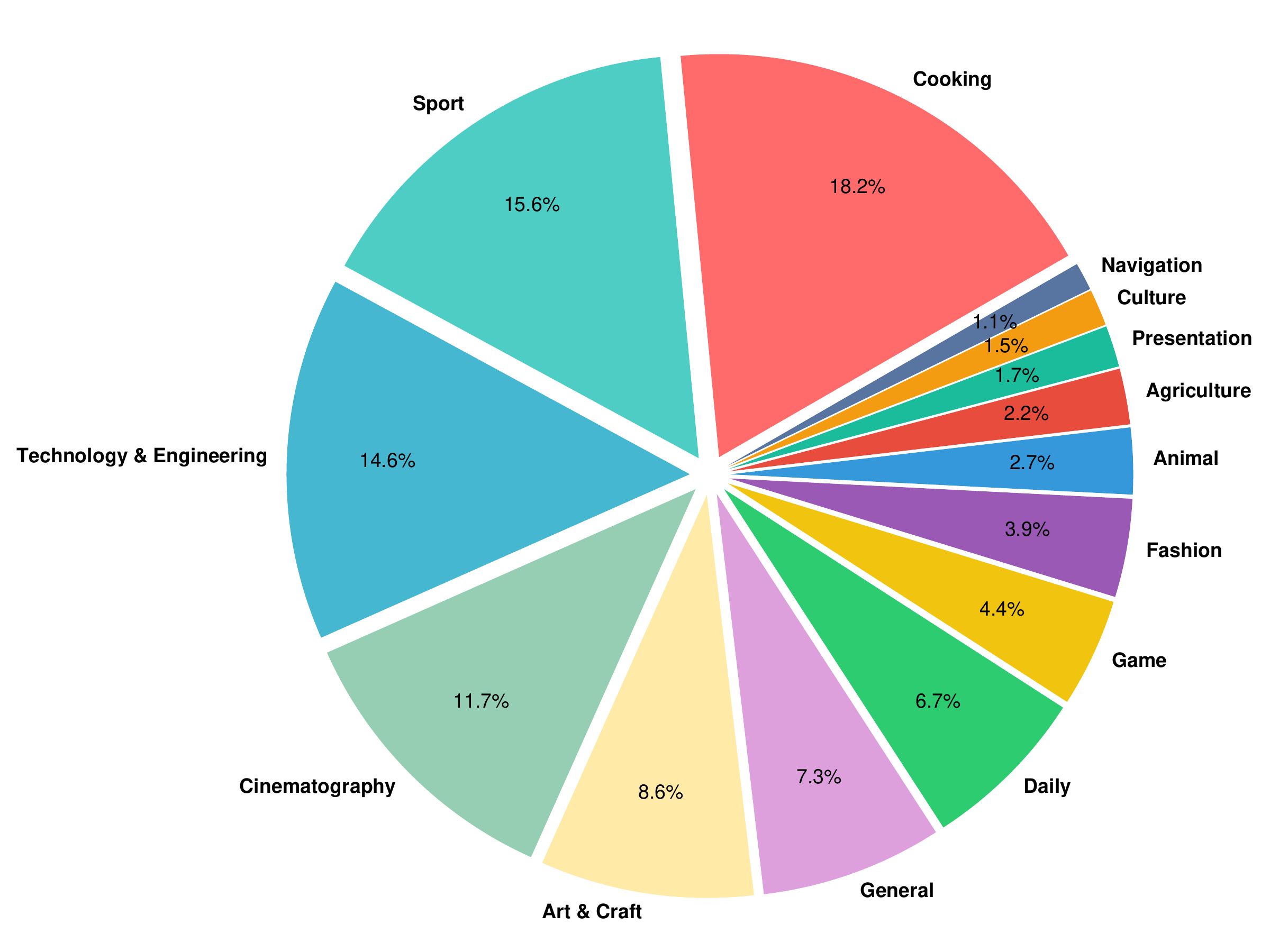}}
  \caption{Topic distribution for TwiFF-2.7M.}
  \label{fig:topicdistribution}
\end{figure}
\section{Dataset Analysis}
\subsection{Data Composition}
Examples of the dynamic VCoT data from TwiFF-2.7M and TwiFF-Bench are illustrated in~\cref{fig:datasetoverview}, encompassing a diverse range of VCoT sequences with varying lengths across three distinct tasks. In TwiFF-2.7M, $70.6\%$ of the tasks are instructional, $17.5\%$ are predictive, and $11.9\%$ pertain to camera. In TwiFF-Bench, $71.2\%$ of the tasks are instructional, $15.6\%$ are predictive, and $13.2\%$ relate to camera. In addition to task categories, we also categorize the knowledge contained in TwiFF-2.7M. Specifically, we input each question-answer pair into Qwen3~\cite{yang2025qwen3} to obtain coarse-grained topic predictions. We then extracted semantic embeddings for the coarse-grained topics using Qwen3-Embedding~\cite{zhang2025qwen3}, applied agglomerative clustering on these embeddings to obtain 40 initial clusters, and subsequently manually merged semantically redundant or overlapping groups, resulting in a final set of 14 distinct topics. The final topic distribution of the dataset is shown in the~\cref{fig:topicdistribution}. TwiFF-2.7M spans 14 distinct topics: Cooking, Sports, Technology \& Engineering, Cinematography, Art \& Craft, General, Daily, Game, Fashion, Animals, Agriculture, Presentation, Culture, and Navigation.

Moreover, TwiFF-2.7M contains VCoTs of diverse lengths. Specifically, approximately $64\%$ of the VCoT instances involve a single image in their reasoning process, $29\%$ contain two images, and the remaining $7\%$ span between three to seven images. Notably, even among samples with the same number of visual cues, the temporal duration represented by each image varies significantly. By computing the time interval between the key frames (including the frame in question) corresponding to the images in each sample within the original video, we estimate the average temporal span of the underlying action. As illustrated in~\cref{fig:frametime}, the majority of samples exhibit action durations under 10 seconds, while the longest extend beyond 40 seconds—highlighting the rich temporal dynamics and diversity of events captured in TwiFF-2.7M.

%---------------------------------------------------------------------------------------
\subsection{Data Quality Assessment}

We leveraged Qwen3VL to evaluate the data quality of TwiFF-2.7M. Specifically, we first randomly sampled 10,000 instances from the dataset. For each instance, the evaluation model was instructed to assess two key criteria: (1) Answerability: Given only the input image and accompanying text, can the model reasonably infer the provided answer? (2) Logical Coherence: Is the reasoning process in the VCoT both logically sound and sufficient to derive the final answer? Based on this assessment, any instance failing to satisfy either criterion was labeled as unacceptable. Remarkably, only $7.3\%$ of the sampled instances were classified as unacceptable, underscoring the exceptionally high quality of the TwiFF-2.7M dataset.
\section{Experiment}
\subsection{Experiment Setup}
\textbf{Training Setup.} We finetune Bagel-7B~\cite{deng2025emerging} on our dataset, using a learning rate of $2 \times 10^{-5}$ with cosine decay. TwiFF is trained for $36,000$ steps on the full TwiFF-2.7M dataset. For ablation studies, TwiFF-Lite, TwiFF-Text, and TwiFF-Image were each trained for $6,000$ steps on the same 300K-sample subset of TwiFF-2.7M. Specifically, we train TwiFF-Text by retaining only the textual components of VCoT, and TwiFF-Image by retaining solely its visual the visual components. To ensure that TwiFF-Image can reliably determine when to terminate the image generation process, we retain the literal text ``frame\_i" from VCoT to maintain the original interleaved text-image data format. The training data for TwiFF-Image follows this structure: ``frame\_i"+ $\langle \textit{frame}_i \rangle$ + ``frame\_i+1" + $\langle \textit{frame}_i \rangle$ + ... + $\langle \text{final answer} \rangle$. (See~\cref{APP:TrainSetup} for details)

\textbf{Evaluation Models.} We evaluated TwiFF and other models on TwiFF-Bench and Seed-Bench-R1. The compared models fall into four categories: (1) understanding-only multimodal models adopting the TCoT paradigm, including Qwen2.5VL-7B~\cite{bai2025qwen2}, Qwen3VL-8B~\cite{bai2025qwen3}, and InternVL3.5-8B~\cite{wang2025internvl3}; (2) unified models adopting the TCoT paradigm such as Bagel-7B and Janus-Pro-7B~\cite{chen2025janus}; (3) understanding-only multimodal models with tool based VCoT, represented by DeepEyes~\cite{zheng2025deepeyes}; and (4) unified models with VCoT, including Zebra-CoT~\cite{li2025zebra} and ThinkMorph~\cite{gu2025thinkmorph}. To prevent the model in VCoT from entering an infinite loop of image generation during the reasoning process and failing to terminate, we impose explicit limits on tool usage and image generation. Specifically, for DeepEyes, we cap the maximum number of tool invocations at 5. For ThinkMorph, Zebra-CoT, and TwiFF, we set the maximum number of image generations to 8. Any response exceeding these respective limits is truncated to prevent infinite loops.

\textbf{Benchmark Setup.} We employ GPT-5.1-2025-11-13 as a judge. On TwiFF-Bench, we evaluate both the CoT score and the answer score of models. On Seed-Bench-R1, due to the absence of reference reasoning chains, we evaluate only the answer score. All scores are bounded between 0 and 5. The specific details of the scoring prompt are provided in the~\cref{APP:evalprompt}. For TwiFF-Bench, we report average scores across all samples as well as separately for three distinct tasks: Instructional, Predictive, and Camera. For Seed-Bench-R1, we present average scores over all samples and across three different scenarios—L1, L2, and L3. Specifically, L1 and L2 consist of first-person egocentric activity scenes centered around kitchen-related tasks, sourced respectively from Epic-Kitchens and Ego4D. In contrast, L3 encompasses a broader range of first-person scenarios from Ego4D beyond the kitchen domain, including activities related to hobbies, recreation, and work.
%-------------------------------------------------------------------------
\subsection{Overall Performance}
From~\Cref{tab:main_results}, we observe the following key findings: \textbf{(1) Dynamic VCoT significantly enhances both the rationality of reasoning processes and the accuracy of answer in dynamic scenarios.} Compared to the base model Bagel, TwiFF achieves a $28.8\%$ improvement in CoT scores and a $41.6\%$ improvement in answer scores on the TwiFF-Bench. Moreover, it demonstrates strong generalization, with a $21.0\%$ improvement in answer scores on the OOD Seed-Bench-R1 dataset. \textbf{(2) Dynamic VCoT outperforms both TCoT and static VCoT on tasks involving dynamic scenarios.} On TwiFF-Bench, our method surpasses existing models employing TCoT-style reasoning as well as various static VCoT variants—including tool-augmented and generative implementations. TwiFF surpasses all competing methods except Qwen3VL-8B, achieving the second-highest performance overall. Since our base model Bagel is built upon Qwen2.5~\cite{yang2024qwen2_5}, this residual gap likely stems from architectural improvements in Qwen3VL. Even with less data, TwiFF-Lite performs competitively, highlighting our framework's efficiency and scalability.

\begin{table*}[t!]
    \caption{\textbf{Main experimental results.} We present the performance of different models on TwiFF-Bench and Seed-Bench-R1. The best-performing model on each benchmark is shown in \textbf{bold}, and the second-best is \underline{underlined}. Qwen3VL refers to the Qwen3VL-Think version.}
    \label{tab:main_results}
    \small
    \centering
    \setlength{\tabcolsep}{3pt}
    \begin{tabular*}{0.98\textwidth}{@{\extracolsep{\fill}}lccccccccccccc@{\hspace{4pt}}}
        \toprule
        \multirow{3}{*}[-6pt]{\textbf{Model}} & 
        \multirow{3}{*}[-6pt]{\textbf{Size}} &
        \multicolumn{8}{c}{\textbf{TwiFF-Bench}} & 
        \multicolumn{4}{c}{\textbf{Seed-Bench-R1}} \\
        \cmidrule(lr){3-10}
        \cmidrule(lr){11-14} 
        & & 
        \multicolumn{2}{c}{\textbf{Instructional}} &
        \multicolumn{2}{c}{\textbf{Predictive}} &
        \multicolumn{2}{c}{\textbf{Camera}} &
        \multicolumn{2}{c}{\textbf{Avg}} &
        \textbf{L1} & \textbf{L2} & \textbf{L3} & \multirow{2}{*}[-4pt]{\textbf{Avg}} \\
        \cmidrule(lr){3-4} \cmidrule(lr){5-6} \cmidrule(lr){7-8}
        \cmidrule(lr){9-10} \cmidrule(lr){11-11} \cmidrule(lr){12-12} \cmidrule(lr){13-13}
        & & \textbf{CoT} & \textbf{Ans} & \textbf{CoT} & \textbf{Ans} & \textbf{CoT} & \textbf{Ans} & \textbf{CoT} & \textbf{Ans} & \textbf{Ans} & \textbf{Ans} & \textbf{Ans} & ~ \\
        \midrule
        \multicolumn{14}{c}{\textit{Understanding-only Multimodal Models}} \\
        \midrule
        Qwen2.5VL & 7B & 2.40 & 1.64 & 2.57 & 1.77 & 2.64 & 1.41 & 2.46 & 1.63 & 1.49 & 1.60 & 1.29 & 1.45 \\
        InternVL3.5 & 8B & 2.24 & 1.76 & 2.50 & 2.27 & 2.77 & 1.83 & 2.35 & 1.85 & 1.58 & \underline{1.67} & 1.39 & 1.55 \\
        Qwen3VL & 8B & \underline{2.78} & 2.31 & 2.88 & 2.80 & 3.10 & 2.76 & 2.84 & 2.44 & \textbf{1.95} & \textbf{1.94} & \textbf{1.65} & \textbf{1.87} \\
        \midrule
        \multicolumn{14}{c}{\textit{Understanding-only Multimodal Models with Tool Based VCoT}} \\
        \midrule
        DeepEyes & 7B & 2.49 & 2.09 & 2.76 & 2.69 & 2.57 & 2.17 & 2.54 & 2.20 & 1.57 & 1.64 & 1.28 & 1.50 \\
        \midrule
        \multicolumn{14}{c}{\textit{Unified Models}} \\
        \midrule
        Janus-Pro & 7B & 1.95 & 1.00 & 2.14 & 1.36 & 2.38 & 0.93 & 2.04 & 1.04 & 1.25 & 1.29 & 0.98 & 1.15 \\
        Bagel & 7B & 2.20 & 1.72 & 2.45 & 2.10 & 2.61 & 2.22 & 2.29 & 1.85 & 1.36 & 1.48 & 1.22 & 1.34 \\
        \midrule
        \multicolumn{14}{c}{\textit{Unified Models with VCoT}} \\
        \midrule
        Zebra-COT & 7B & 2.16 & 1.22 & 2.61 & 2.10 & 2.47 & 1.63 & 2.27 & 1.41 & 1.09 & 1.19 & 1.11 & 1.11 \\
        ThinkMorph & 7B & 2.15 & 1.30 & 2.48 & 2.38 & 2.20 & 1.00 & 2.21 & 1.43 & 1.07 & 1.13 & 1.03 & 1.07 \\
        TwiFF-Lite & 7B & 2.76 & \underline{2.34} & \underline{3.13} & \underline{3.03} & \textbf{3.34} & \underline{3.11} & \underline{2.90} & \underline{2.55} & 1.56 & 1.62 & 1.50 & 1.55 \\
        \textbf{TwiFF} & 7B & \textbf{2.81} & \textbf{2.43} & \textbf{3.24} & \textbf{3.04} & \underline{3.32} & \textbf{3.14} & \textbf{2.95} & \textbf{2.62} & \underline{1.64} & \underline{1.67} & \underline{1.56} & \underline{1.62} \\
        \midrule
        \textcolor{softgreen}{Gain(\%) (vs BAGEL)} & 
        - & \textcolor{softgreen}{\textbf{+27.7}} &
        \textcolor{softgreen}{\textbf{+41.3}} & \textcolor{softgreen}{\textbf{+32.2}} &
        \textcolor{softgreen}{\textbf{+44.8}} & \textcolor{softgreen}{\textbf{+27.2}} &
        \textcolor{softgreen}{\textbf{+41.4}} &
        \textcolor{softgreen}{\textbf{+28.8}} & \textcolor{softgreen}{\textbf{+41.6}} & \textcolor{softgreen}{\textbf{+20.6}} & \textcolor{softgreen}{\textbf{+12.8}} & \textcolor{softgreen}{\textbf{+27.9}} & \textcolor{softgreen}{\textbf{+21.0}} \\
        \bottomrule
    \end{tabular*}
\end{table*}
%-------------------------------------------------------------------------%
\label{sec:modabl}
\begin{table}
  \caption{Ablation study on the effectiveness of visual and textual modalities in VCoT.}
  \label{tab:ablationVCoT}
  \small
  \centering
  \begin{tabular*}{0.48\textwidth}{@{\extracolsep{\fill}}lcccc@{\hspace{4pt}}}
    \toprule
    \multirow{2}{*}[-4pt]{\textbf{Method}} & 
    \multicolumn{3}{c}{\textbf{TwiFF-Bench}} & 
    \textbf{Seed-Bench-R1} \\
    \cmidrule(lr){2-4} \cmidrule(lr){5-5}
     & \textbf{CoT} & \textbf{Ans} & \textbf{Avg} & \textbf{Avg} \\
    \midrule
    Bagel & 2.29 & 1.85 & 2.07 & 1.34 \\
    TwiFF-Text & 2.80 & 2.47 & 2.64 & 1.46 \\
    TwiFF-Image & 2.25 & 2.50 & 2.38 & 1.37 \\
    TwiFF-Lite & \textbf{2.90} & \textbf{2.55} & \textbf{2.73}  & \textbf{1.62} \\
    \bottomrule
  \end{tabular*}
\end{table}
\subsection{Modality Ablation in VCoT}
In this section, we conduct ablation studies to analyze the necessity and effectiveness of each modality-specific reasoning cue in VCoT. TwiFF-Text relies exclusively on textual cues during inference, TwiFF-Image solely on visual cues, and TwiFF-Lite interleaves the generation of textual and visual cues. As shown in \cref{tab:ablationVCoT}, we can draw the following conclusions: \textbf{(1) Interleaved vision-textual VCoT demonstrates superior reasoning and generalization capabilities compared to single-modality CoTs.} TwiFF-Lite outperforms TwiFF-Text by $3.4\%$ and TwiFF-Image by $14.7\%$ on the TwiFF-Bench. On the OOD benchmark Seed-Bench-R1, these increases are even more pronounced at $11.0\%$ and $18.2\%$, respectively. This clearly indicates that the enhanced reasoning abilities of TwiFF-Lite stem from the synergistic interaction between textual and visual modalities. \textbf{(2) CoT reasoning within a single modality exhibits limited generalization capability on OOD benchmark.} Compared to Bagel, while TwiFF-Text and TwiFF-Image achieve substantial improvements in answer accuracy on TwiFF-Bench, their performance gains on Seed-Bench-R1 are modest—only $9.0\%$ and $2.2\%$, respectively. In contrast, TwiFF-Lite, which leverages visual-textual reasoning, demonstrates a significantly higher improvement of $20.9\%$. Thus, the generalization capability of TwiFF-Lite in dynamic scenes stems from the interplay between visual and textual cues.
%-------------------------------------------------------------------------%
\begin{figure}[t!]
  \centering
  \centerline{\includegraphics[width=0.95\linewidth]{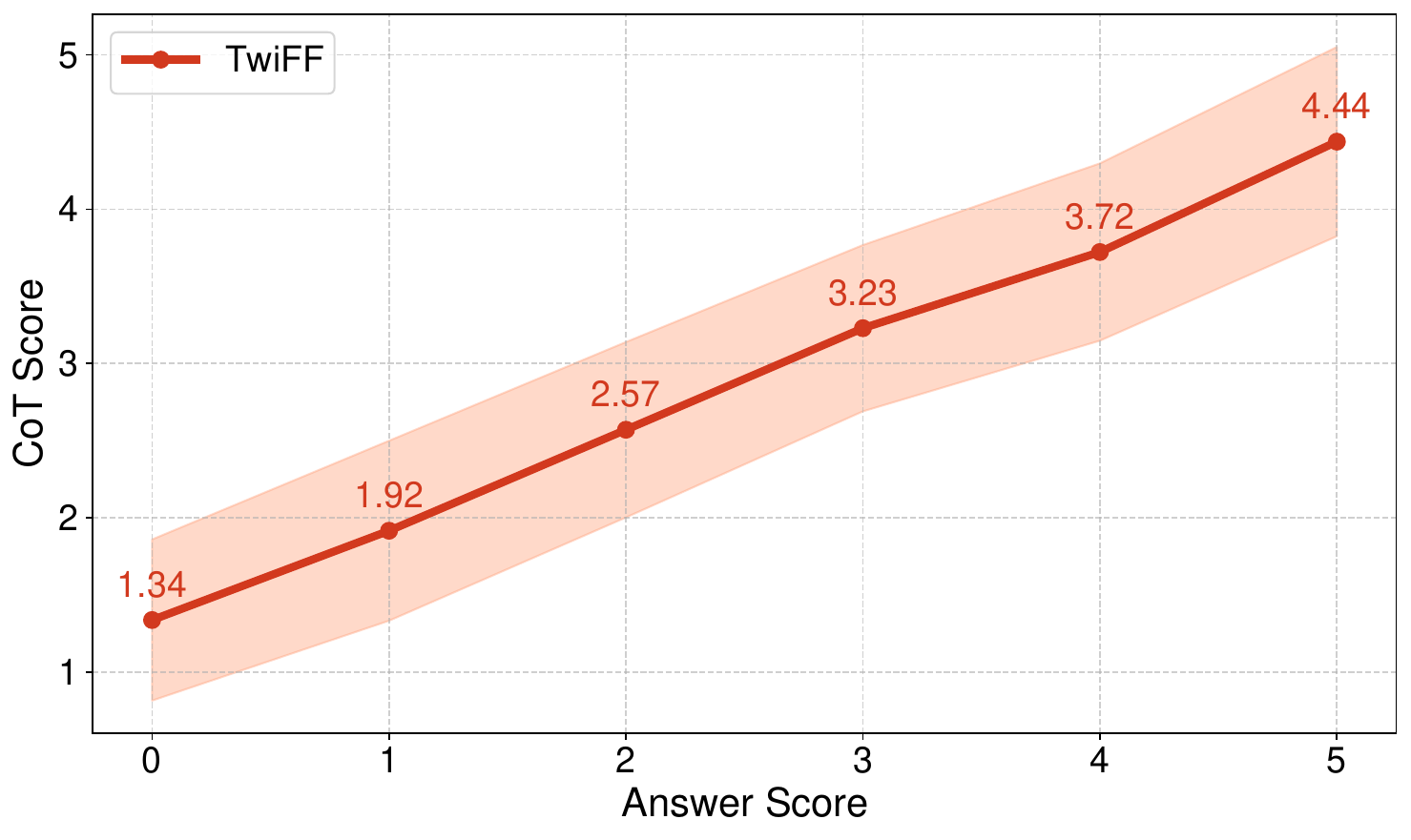}}
  \caption{The impact of the alignment between VCoT reasoning and observed future events in the video on the answer score on TwiFF-Bench.}
  \label{fig:cot-ans}
\end{figure}
\begin{table}[t!]
  \caption{\textbf{Impact of Visual Cues.} On TwiFF-Bench, we evaluate how the quality of visual cues affects model answer scores.}
  \label{tab:VCoTquality}
  \small
  \centering
  \begin{tabular}{@{\extracolsep{\fill}}lccc@{}}
    \toprule
    \textbf{Method} & 
    \textbf{CoT} &
    \textbf{Ans} \\
    \midrule
    TwiFF & 2.95 & 2.62 & \\
    TwiFF-True & 3.56 (\textcolor{softgreen}{$20.7\% \uparrow$}) & 3.14 (\textcolor{softgreen}{$19.8\% \uparrow$}) & \\
    TwiFF-False & 2.92 (\textcolor{softred}{$1.0\% \downarrow$}) & 2.57 (\textcolor{softred}{$1.9\% \downarrow$}) & \\
    \bottomrule
  \end{tabular}
\end{table}
\subsection{The Impact of Dynamic VCoT Quality on the Score of Model Answer}
In this section, we analyze the results from TwiFF-Bench to quantify how the quality of dynamic VCoT influences model's final answer. Since the reference VCoTs are grounded in observed future events in the video, a higher CoT score indicates stronger alignment between the model's generated reasoning and the actual subsequent events. As shown in Figure~\ref{fig:cot-ans}, higher-quality VCoTs consistently correlate with higher answer scores, demonstrating that answer accuracy critically depends on how well the reasoning matches reality.

Furthermore, we conduct controlled experiments to analyze how the dynamism and quality of visual cues in the TwiFF reasoning process affect model performance. During inference, for responses generated by the TwiFF model, we truncate the VCoT at the point where the first image is generated.
In TwiFF-True, we replace the model's generated initial image with the ground-truth first image from the reference VCoT in TwiFF-Bench, providing a visual cue aligned with actual future events. Conversely, in TwiFF-False, we replace it with a duplicate of the input question image, simulating a failure mode in which the model refrains from predicting dynamic future states and instead produces a static, uninformative visual cue. Both variants then continue generating the remainder of the VCoT until producing the final answer. However, because the model may still generate informative visual content in later steps of the reasoning process, TwiFF-False does not fully restrict it to static visual reasoning.
As shown in Table~\ref{tab:VCoTquality}, TwiFF-True achieves significant improvements over the original TwiFF model in both cot score and answer score, whereas TwiFF-False exhibits degradation in both metrics. These results underscore the critical importance of temporal fidelity and semantic correctness in visual reasoning cues: when the visual cues faithfully reflect the true dynamics of future events (as in TwiFF-True), model performance is substantially enhanced; in contrast, static or misleading cues that lack dynamic content (as in TwiFF-False) actively mislead the reasoning process and degrade answer correctness.
\begin{table}[t!]
  \caption{Exploring the potential of information compression in VCoT: On TwiFF-Bench, we evaluate the information compression capability of visual cues in VCoT by discarding the input question image.}
  \label{tab:VCoTCompress}
  \small
  \centering
  \begin{tabular}{@{\extracolsep{\fill}}lccc@{}}
    \toprule
    \textbf{Method} & 
    \textbf{CoT} &
    \textbf{Ans} \\
    \midrule
    TwiFF & 2.95 & 2.62 & \\
    TwiFF-Comp & 2.87 (\textcolor{softred}{$2.7\% \downarrow$}) & 2.50 (\textcolor{softred}{$4.6\% \downarrow$})& \\
    TwiFF-Drop & 2.23 (\textcolor{softred}{$24.4\% \downarrow$}) & 2.25 (\textcolor{softred}{$14.1\% \downarrow$}) & \\
    \bottomrule
  \end{tabular}
\end{table}
%-------------------------------------------------------------------------%
\subsection{The Potential of Information Compression in VCoT}
In addition to improved reasoning performance and answer plausibility in dynamic scenarios, dynamic VCoT also demonstrates notable potential for information compression. To validate the capacity of VCoT to retain visual cues and compress contextual information, we conduct an ablation study by discarding the input question image at the point when TwiFF generates its first image during inference. Specifically, in the TwiFF-Comp variant, only the original input question image is discarded after this point, and the model proceeds to generate its response using the retained visual representation. In contrast, the TwiFF-Drop variant discards both the input question image and TwiFF's first generated image; all subsequent reasoning steps and final answers are then produced exclusively from textual context.

As shown in Table~\ref{tab:VCoTCompress}, TwiFF-Drop suffers substantial performance degradation, with CoT and answer scores declining by $24.4\%$ and $14.1\%$, respectively. In contrast, TwiFF-Comp exhibits only minor declines—$2.7\%$ for CoT and $4.6\%$ for answer score—indicating that the dynamic VCoT mechanism effectively preserves the contextual information originally in the input image. These results underscore the potential of our approach to enable efficient information compression while mitigating catastrophic forgetting in multi-step reasoning with visual cues.
\section{Conclusion}
\label{sec:conclusion}
In this work, we aim to move VCoT beyond the confines of static analysis of input images. To this end, we introduced TwiFF-2.7M, the first large-scale, dynamic VCoT dataset, comprising over 2.7 million video clips annotated with fine-grained, step-by-step visual-textual reasoning that trace both instructional logic and predictive trajectories. Complementing this resource, we proposed TwiFF-Bench, a rigorously curated evaluation benchmark that uniquely assesses both the plausibility of reasoning steps and the correctness of final predictions in dynamic, open-ended scenarios—moving beyond the narrow focus on end-task accuracy that characterizes prior benchmarks.
Our experiment results demonstrate that unified models trained on TwiFF-2.7M achieve substantial gains over existing static VCoT and TCoT baselines in dynamic visual reasoning tasks. Looking ahead, we believe TwiFF-2.7M and TwiFF-Bench will serve as foundational resources for advancing research in dynamic visual reasoning. Moreover, TwiFF reveals the impact of the authenticity of visual cues on answer accuracy, offering a promising direction for the community to explore novel reinforcement learning approaches in multimodal reasoning scenarios—specifically through the lens of CoT plausibility.

\section*{Impact Statement}
This paper presents work whose goal is to advance the field of Machine Learning. There are many potential societal consequences of our work, none which we feel must be specifically highlighted here.

\bibliography{example_paper}
\bibliographystyle{icml2026}

%%%%%%%%%%%%%%%%%%%%%%%%%%%%%%%%%%%%%%%%%%%%%%%%%%%%%%%%%%%%%%%%%%%%%%%%%%%%%%%
%%%%%%%%%%%%%%%%%%%%%%%%%%%%%%%%%%%%%%%%%%%%%%%%%%%%%%%%%%%%%%%%%%%%%%%%%%%%%%%
% APPENDIX
%%%%%%%%%%%%%%%%%%%%%%%%%%%%%%%%%%%%%%%%%%%%%%%%%%%%%%%%%%%%%%%%%%%%%%%%%%%%%%%
%%%%%%%%%%%%%%%%%%%%%%%%%%%%%%%%%%%%%%%%%%%%%%%%%%%%%%%%%%%%%%%%%%%%%%%%%%%%%%%
\newpage
\appendix
\onecolumn
\section{Data Statistics}
\begin{figure*}[h]
  \centering
  \begin{subfigure}{0.49\linewidth}
    \includegraphics[width=\linewidth]{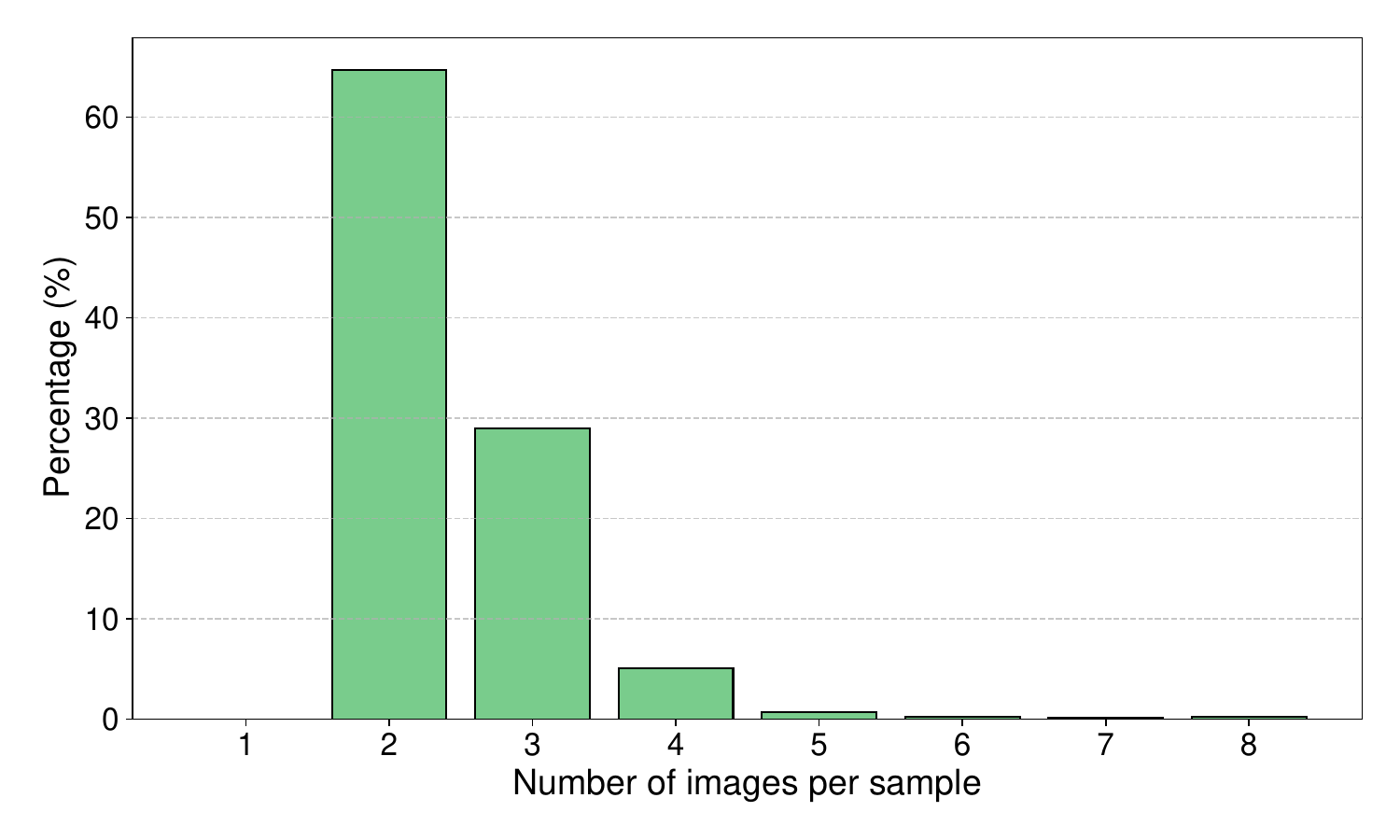}
    \caption{Key frame count per sample}
    \label{fig:framecount}
  \end{subfigure}
  \hfill
  \begin{subfigure}{0.49\linewidth}
    \includegraphics[width=\linewidth]{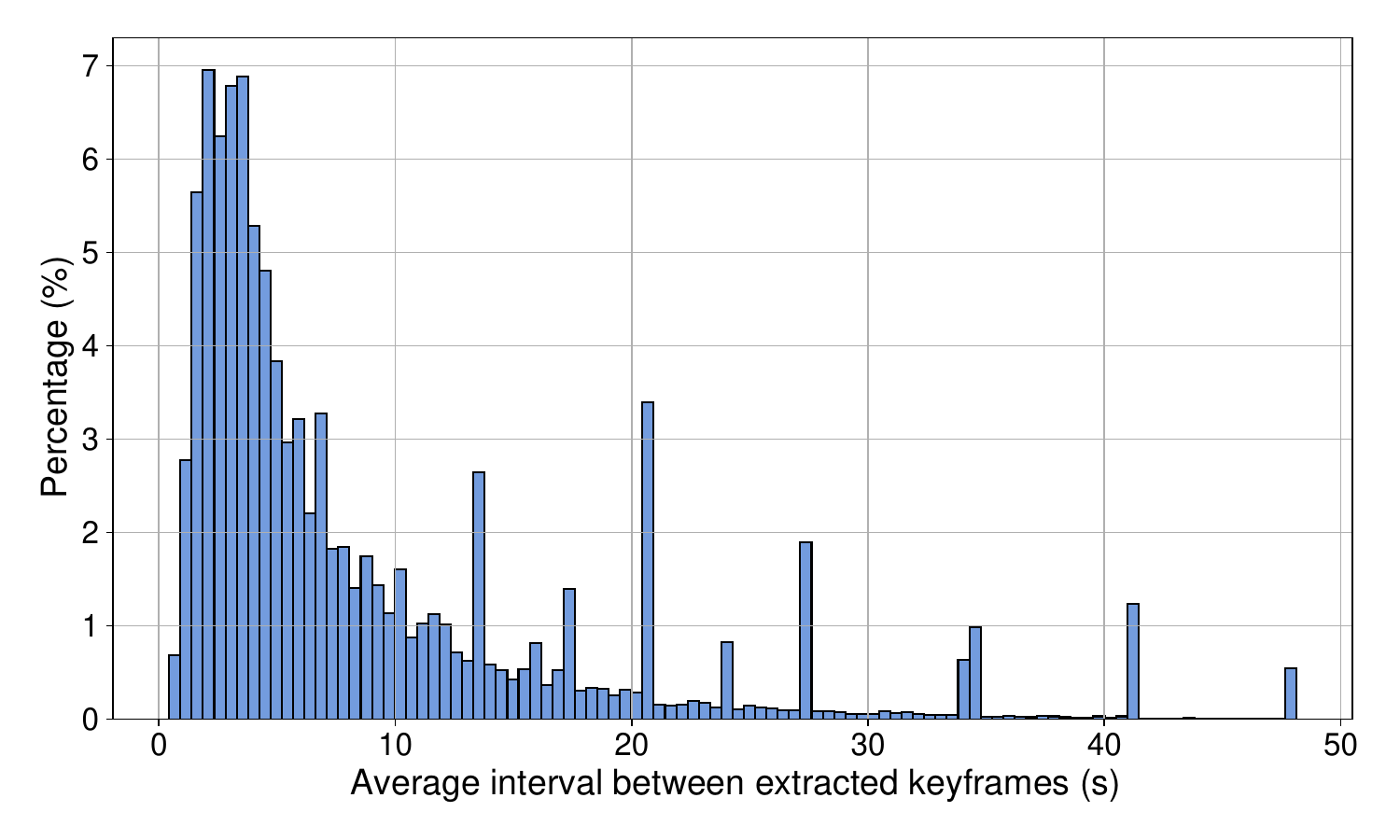}
    \caption{Distribution of average key frame time gaps across samples}
    \label{fig:frametime}
  \end{subfigure}
  \caption{TwiFF-2.7M key frames distribution.}
  \label{fig:data_cnt}
\end{figure*}

\begin{figure*}[h]
  \centering
  \begin{subfigure}{0.49\linewidth}
    \includegraphics[width=\linewidth]{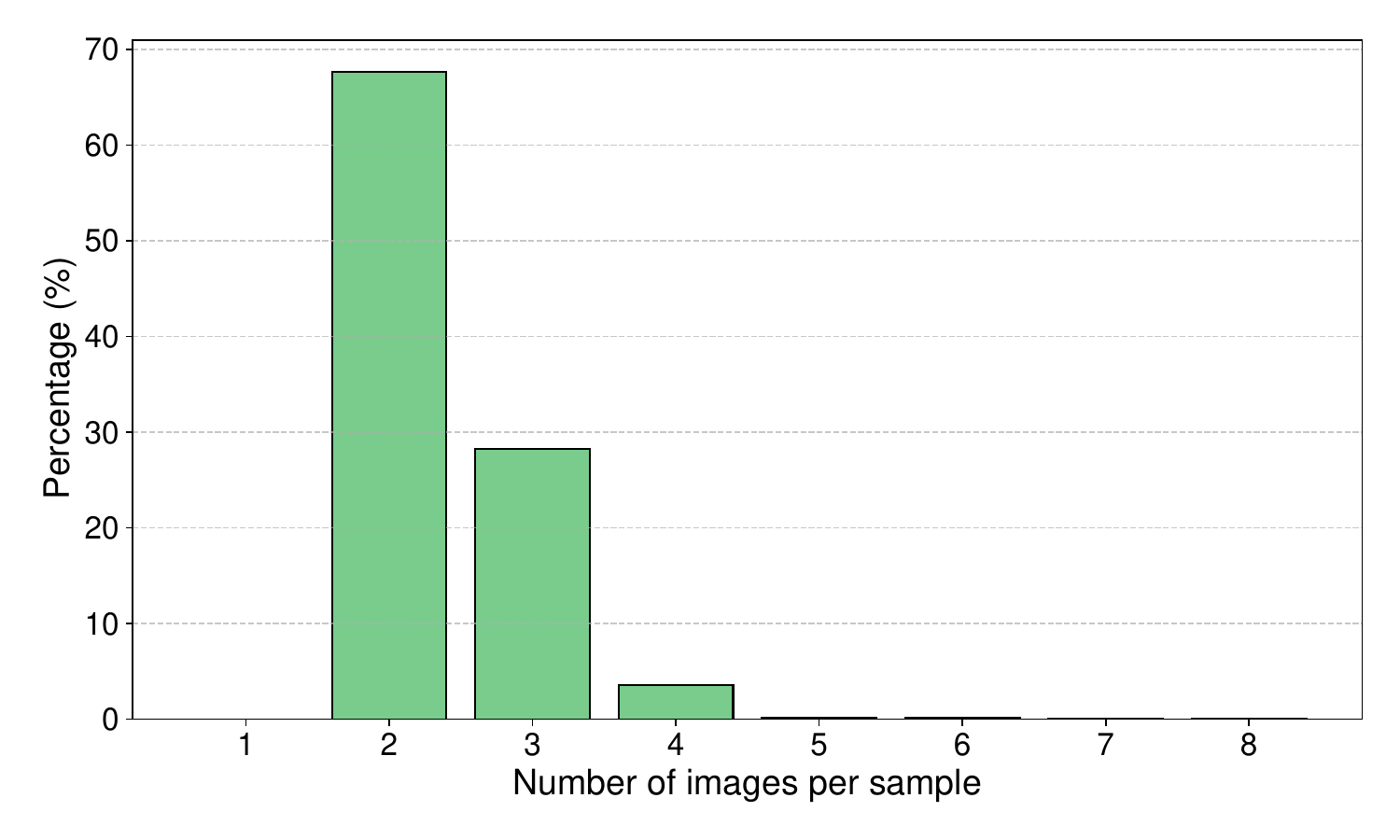}
    \caption{Key frame count per sample}
    \label{fig:benchframecount}
  \end{subfigure}
  \hfill
  \begin{subfigure}{0.49\linewidth}
    \includegraphics[width=\linewidth]{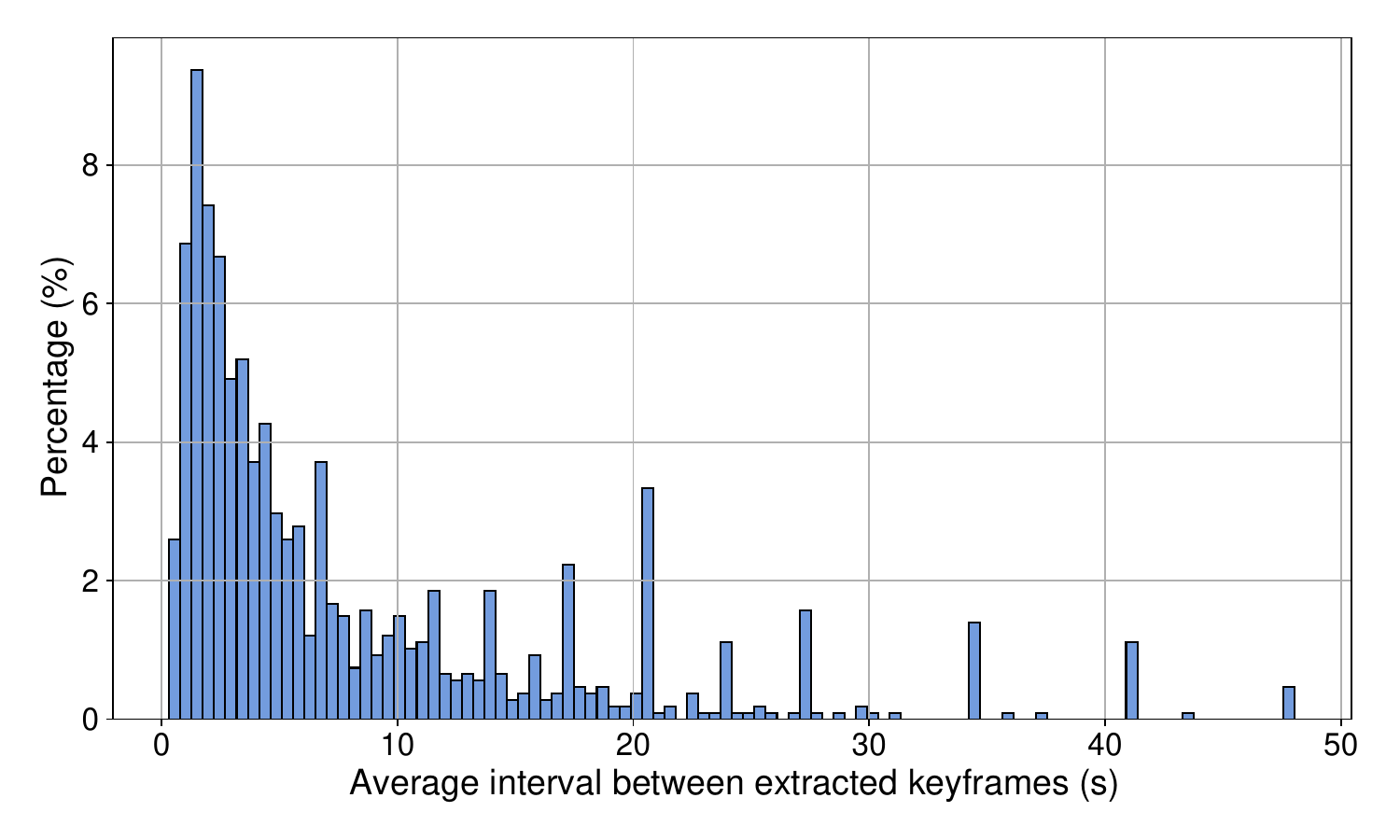}
    \caption{Distribution of average key frame time gaps across samples}
    \label{fig:benchframetime}
  \end{subfigure}
  \caption{TwiFF-Bench key frames distribution.}
  \label{fig:benchdata_cnt}
\end{figure*}
%---------------------------------------------------------------------------%
\section{Prompt Setup}
\subsection{Data Generation Prompt}
\label{APP:dataprompt}
\begin{tcolorbox}[breakable, colback=gray!5!white, colframe=gray!75!black, 
title=Stage 2 Event extraction, boxrule=0.5mm, width=\textwidth, arc=3mm, auto outer arc]
Your task is to analyze a sequence of video frames (provided as a numbered list of image frames in chronological order) along with the video title. All responses must be based solely on the visual content shown in the input frames and adhere to the following instructions:

1. Classify the video into one of four categories: \\
    Instructional: \\
        The video demonstrates a clear, step-by-step procedure or technique through observable, goal-directed physical actions (e.g., cooking, assembling furniture, performing an exercise, repairing a device). The sequence must be visually coherent, imitable, and primarily conveyed through demonstration rather than narration. \\
    Predictive: \\
        The video shows a causal chain grounded in real-world physics or involuntary human behavior, where early frames allow reasonable prediction of later outcomes (e.g., a tower tipping $\rightarrow$ collapsing; a ball rolling toward an edge $\rightarrow$ falling; someone stepping on a banana peel $\rightarrow$ slipping). The progression must be temporally smooth, physically plausible, and feature clear visual dynamics---not just narration or speculation. \\
    Camera: \\
        The video is distinguished by intentional, high-quality dynamic camera work---such as tracking shots, smooth pans, dollies, crane movements, or other deliberate motion that actively shapes the viewer's perception. Static or handheld footage without artistic or functional intent does not qualify. The camera movement itself should be a salient visual element, contributing meaningfully to the scene. \\
    Undesirable: \\
        The video fails to meet the criteria of the above three categories. This includes:
        Videos dominated by speech, text, or narration without substantive visual action;
        Static, blurry, or low-information scenes where the main subject or action is unclear;
        Random cuts, abrupt transitions ($\geq$3 scene changes), or repetitive loops;
        Content too sparse, slow, or abstract to support meaningful visual understanding or QA generation;
        Any video lacking discernible motion, transformation, or intentional visual dynamics. \\
    If a video (not Undesirable) could belong to multiple categories, select one according to the following priority order: Predictive, Instructional, Camera.
    
2. If and only if the video is not classified as ``Undesirable'', select at least two (or more) representative frames. Selected frames must: \\
        Be numbered according to the input list. \\
        Exclude frames with heavy motion blur, digital overlays, transition effects, or other visual artifacts that obscure content clarity. \\
        The number of selected frames should reflect the complexity and steps of the action: simple actions may only need two frames (before + after), while multi-stage events may require additional intermediate frames to illustrate key transitions.

3. If and only if the video is not classified as ``Undesirable'', generate a concise, coherent natural language description across the representative\_frames chosen in steps 2:

    Process: Detail the observable changes, actions, or camera techniques that occur across the \textbf{representative frames}, focus on just one narrative subject. \\
        For Instructional: Emphasize clear, step-by-step actions or demonstrations that convey what to do. \\
        For Predictive: Highlight visual cues (e.g., motion, deformation, environmental shifts) that logically suggest an imminent outcome. \\
        For Camera: Specify camera movements (e.g., dolly, pan, handheld shake, zoom), shot transitions (e.g., cut, fade, match cut), or compositional choices (e.g., rule of thirds, shallow depth of field).

    Summary: Concisely encapsulate the core takeaway: \\
        In instructional videos, explain the logical rationale behind the procedure or skill being taught. \\
        In Predictive videos, state the event result. \\
        In Camera videos, infer or summarize the purpose or effect of this filming technique.
        Note: All descriptions must be grounded solely in what is directly visible in the representative frames---no external assumptions or speculative content.

4. Output your response \textbf{strictly in JSON format} with the following structure: \\
   \{ \\
     ``classification'': ``Instructional'' \textbar\ ``Predictive'' \textbar\ ``Camera'' \textbar\ ``Undesirable'', \\
     ``representative\_frames'': [int]  \# Only include if classification is not ``Undesirable'', \\
     ``Process'': str, \\
     ``Summary'': str, \\
   \}

Do not include any text outside the JSON object. Ensure all frame numbers in ``representative\_frames'' are present in the input.
\end{tcolorbox}

\begin{tcolorbox}[breakable, colback=gray!5!white, colframe=gray!75!black, 
title=Stage 3 VCoT generation, boxrule=0.5mm, width=\textwidth, arc=3mm, auto outer arc]
You are given the following information about a short video clip:

Classification: the type of the video

Process Description: what happened in the video

Summary:

    In instructional videos, explain the logical rationale behind the procedure or skill being taught.

    In Predictive videos, state the event result.

    In Camera videos, infer or summarize the purpose or effect of this filming technique.

Additionally, you will receive a sequence of representative frames as input images.

Your task is to generate one question-answer pair that reflects a coherent, visually grounded reasoning process, with the following requirements:

[Question]

Ask a natural, focused question based solely on what is visible in frame\_1. The question must not be answerable from frame\_1 alone; it should require new information that appears only in later frames (frame\_i, i \textgreater 1). frame\_1 appears in the question and is not part of the reasoning Chain. Do not assume or refer to content from later frames unless it's introduced as an explicit conditional (e.g., ``If the person in frame\_1\dots'') within the question itself.

The question must also align with the video's classification:

If Instructional, ask how to perform a task, what step should follow next, or what actions are necessary to achieve a specific goal-with the required procedural details revealed in the later frames.

If Predictive, ask whether a particular outcome will occur or what result will follow if a condition holds-where the confirmation or realization of that outcome unfolds across later frames.

In Camera, ask: How should the camera move, frame, or compose the shot to emphasize a specific element, convey meaning, or evoke a particular emotion-so that the expressive intent only becomes clear as the shot unfolds.

[Answer]

The answer consists of a reasoning chain and the final answer
Reasoning Chain (including frame\_i, i\textgreater=2):

In a single, fluent paragraph, explain how the answer is progressively supported by observations from each of frame\_2 through frame\_n, referenced exactly once each (e.g., ``frame\_i shows'', ``By frame\_i'', ``In frame\_i'', etc.). Use natural transitions (``then'', ``as it continues'', ``shortly after'') to maintain narrative flow-never skip, duplicate, or reorder frames. Every claim must be directly verifiable from the visuals and consistent with the video's overall process or intent. The reasoning chain should follow the format below:

\textless text description\textgreater\ + \textless frame\_i\textgreater\ + \textless reasoning about i\textgreater\ + \textless frame\_i+1\textgreater\ + \textless reasoning about i+1\textgreater+\dots

Final Answer:

End with a concise, definitive response to the question, enclosed strictly between \textless ans\textgreater\ and \textless/ans\textgreater.

The final answer must synthesize only the observable content and logical progression described in the reasoning chain, without any mention of frames, frame numbers.

Style Guidelines:

Write in a natural, human-like voice. Avoid formulaic repetition. Match the reasoning style to the classification:

Instructional: Clarify what is done and why, step by step, as revealed over time.\\
Predictive: Show how early signs or actions inevitably lead to a later outcome.\\
Camera: Trace how evolving camera choices shape interpretation or feeling across the sequence.\\
Output Format (plain text):

[Question]\{Question with information, across frame\_1\}

[Answer]

\{Step-by-step visual reasoning across frame\_2 to frame\_n\dots\}
\textless ans\textgreater\{Final concise answer\}\textless/ans\textgreater

\end{tcolorbox}
%---------------------------------------------------------------------------%
\subsection{GPT Evaluation Prompt}
\label{APP:evalprompt}
\begin{tcolorbox}[breakable, colback=gray!5!white, colframe=gray!75!black, 
title=TwiFF-Bench Evaluation Prompt, boxrule=0.5mm, width=\textwidth, arc=3mm, auto outer arc]
You are a strict evaluator. You will have to evaluate the model response reasoning chain and answer based on the reference reasoning chain and ground truth answer.\\
Given:\\
    Question: The original forecasting question with image originates from the first video frame.\\
    Reference Reasoning Chain: What actually happened, as a reference for the rationality of the reasoning chain.\\
    Ground Truth Answer: The ground truth of the question.\\
    Model Response Reasoning Chain: The model's reasoning chain.\\
    Model Response Answer: The model's answer.\\
The rating should base on the following rules:\\
    Reasoning Chain Quality: Score 0-5 based on the logical coherence, completeness, and relevance of the reasoning (including appropriate use of multimodal information if present). The chain need not match the reference exactly but must be valid and support the final answer.\\
    Answer Accuracy: Score 0-5 based on how well the final answer matches the ground truth answer. Full credit requires correctness and completeness; partial or incorrect answers receive lower scores.\\
Put the score in a list such that output score = [score1, score2], where 'score1' evaluates the
Reasoning Chain and 'score2' evaluates the Answer.\\
You will have to give your output in the JSON format (Keep your reasoning concise and short.):\\
\{\\
"reasoning": str \#the score reasoning\\
"score": List[int]\\
\}
\end{tcolorbox}

\begin{tcolorbox}[
  breakable,
  colback=gray!5!white,
  colframe=gray!75!black,
  title=SeedBenchR1 Evaluation Prompt,
  boxrule=0.5mm,
  width=\textwidth,
  arc=3mm,
  auto outer arc
]
You are a evaluator. You will have to evaluate the model answer based on the question and ground truth answer.\\
Given:\\
    Question: The original forecasting question with image originates from the first video frame.\\
    Ground Truth Answer: The ground truth of the question.
    Model Response Reasoning Chain: The model's reasoning chain. Some models may generate images as part of their reasoning chain. \\
    Model Response Answer: The model's answer.\\
The rating should base on the following rules:\\
    Answer Accuracy: Score 0-5 based on how well the final answer matches the question requirement and ground truth answer. Full credit requires correctness and completeness; partial or incorrect answers receive lower scores. For answers that differ from the ground truth, appropriately evaluate their reasonableness and assign a score, rather than treating them as entirely incorrect.\\
Put the score in a list such that output score = [score1], where 'score1' evaluates the Answer.\\
You will have to give your output in the JSON format (Keep your reasoning concise and short.):\\
\{{\\
"reasoning": str \#the score reasoning\\
"score": List[int]\\
\}}
\end{tcolorbox}

%---------------------------------------------------------------------------%
\subsection{Inference Prompt}
\begin{tcolorbox}[
  breakable,
  colback=gray!5!white,
  colframe=gray!75!black,
  title=TwiFF System Prompt,
  boxrule=0.5mm,
  width=\textwidth,
  arc=3mm,
  auto outer arc
]
You are an AI assistant capable of reasoning with visual imagery. You should conduct a detailed analysis of the question. Consider different angles, potential solutions, and reason through the problem step-by-step with image. After fully reasoning through the problem—potentially using image-based thinking—provide only a clear, concise, and direct answer to the user's question.
\end{tcolorbox}
%---------------------------------------------------------------------------%
\section{Training and Inference Steup.}
\label{APP:TrainSetup}
The hyperparameters used to train TwiFF are provided in~\cref{tab:trainHyp}. In TwiFF, the text from interleaved image-text sequences is partitioned into multiple segments, separated by images. During training, each text segment and each image is randomly dropped with a certain probability to facilitate Classifier-Free Guidance(CFG) for image generation. However, to ensure that the trained model retains the capability to produce a coherent final output, we never drop the last text segment or the last image during training.

During inference, the model alternately generates text and images until the generated text contains an answer enclosed within \textless ans\textgreater\ and \textless/ans\textgreater\ tags. For TwiFF-Image, TwiFF-Lite, and TwiFF, we set \texttt{temperature=0.3}, \texttt{cfg\_text\_scale=3.5}, \texttt{cfg\_img\_scale=2.0} for image generation, and \texttt{max\_tokens=4,096} for text generation in each interleaved segment. Notably, since TwiFF-Text produces only textual responses, it adopts the same inference configuration as Bagel: \texttt{temperature=0.0} and \texttt{max\_tokens=8,192}.

\begin{table}[h]
  \caption{\textbf{Training Hyperparameters.} ``-'' denotes that the parameter is not applicable.}
  \label{tab:trainHyp}
  \small
  \centering
  \begin{tabular}{@{\extracolsep{\fill}}lcccc@{}}
    \toprule
    Hyperparameter & 
    \textbf{TwiFF-Text} &
    \textbf{TwiFF-Image} & 
    \textbf{TwiFF-Lite} &
    \textbf{TwiFF} \\
    \midrule
    Dataset Size & 300,000 & 300,000 & 300,000 & 2,708,318 \\
    Max Learning Rate & $2 \times 10^{-5}$ & $2 \times 10^{-5}$ & $2 \times 10^{-5}$ & $2 \times 10^{-5}$ \\
    Min Learning Rate & $1 \times 10^{-6}$ & $1 \times 10^{-6}$ & $1 \times 10^{-6}$ & $1 \times 10^{-6}$ \\
    Learning Rate Scheduler & Cosine Decay & Cosine Decay & Cosine Decay & Cosine Decay \\
    Training Steps & 6,000 & 6,000 & 6,000 & 36,000 \\
    CE Loss Weight & 1.0 & 1.0 & 1.0 & 1.0 \\
    MSE Loss Weight & - & 1.0 & 1.0 & 1.0 \\
    Frozen Components & Generation Expert & None & None & None \\
    Max Tokens per Batch & 10,240 & 36,864 & 36,864 & 36,864 \\
    Text Condition Drop & - & 0.1 & 0.1 & 0.1 \\
    ViT Condition Drop & - & 0.3 & 0.3 & 0.3 \\
    VAE Condition Drop & - & 0.3 & 0.3 & 0.3 \\
    ViT Image Size ([min, max]) & [256, 512] & [256, 512] & [256, 512] & [256, 512] \\
    VAE Image Size ([min, max]) & [224, 518] & [224, 518] & [224, 518] & [224, 518] \\
    \bottomrule
  \end{tabular}
\end{table}
%---------------------------------------------------------------------------%
\section{Inference Token Cost.}
We compute the average token length of model-generated responses on TwiFF-Bench as a proxy for estimating inference overhead. For computational convenience, all images are preprocessed into their corresponding VAE and ViT tokens. The results are summarized in~\cref{tab:TokenCost}. Notably, the average number of images generated per response for TwiFF, TwiFF-Lite, and TwiFF-Image is 1.23, 1.18, and 1.50, respectively.
\begin{table}[h]
  \caption{Inference token cost.}
  \label{tab:TokenCost}
  \small
  \centering
  \begin{tabular}{@{\extracolsep{\fill}}lc@{}}
    \toprule
    Model & 
    Token Cost \\
    \midrule
    Bagel & 1422.40 \\
    TwiFF-Text & 176.08 \\
    TwiFF-Image & 1414.50 \\
    TwiFF-Lite & 1256.45 \\
    TwiFF & 1283.11 \\
    \bottomrule
  \end{tabular}
\end{table}
%---------------------------------------------------------------------------%
\section{Optical Flow Computation}
\label{APP:OpticalFlowComp}
We compute dense optical flow using the \texttt{cv2.calcOpticalFlowFarneback} function from OpenCV, with parameters specified in Table~\ref{tab:OpticalFlow}.
\begin{table}[h]
  \caption{Optical flow parameters.}
  \label{tab:OpticalFlow}
  \small
  \centering
  \begin{tabular}{@{\extracolsep{\fill}}l|c|l|c@{}}
    \toprule
    \textbf{Parameters} & 
    \textbf{Value} &
    \textbf{Parameters} & 
    \textbf{Value} \\
    \midrule
    pyr\_scale & 0.5 & poly\_n & 5 \\
    levels & 3 & poly\_sigma & 1.2 \\
    winsize & 15 & iterations & 3 \\
    \bottomrule
  \end{tabular}
\end{table}

\newpage
\section{Case Study}
\begin{figure}[h]
  \centering
  \centerline{\includegraphics[width=0.95\linewidth]{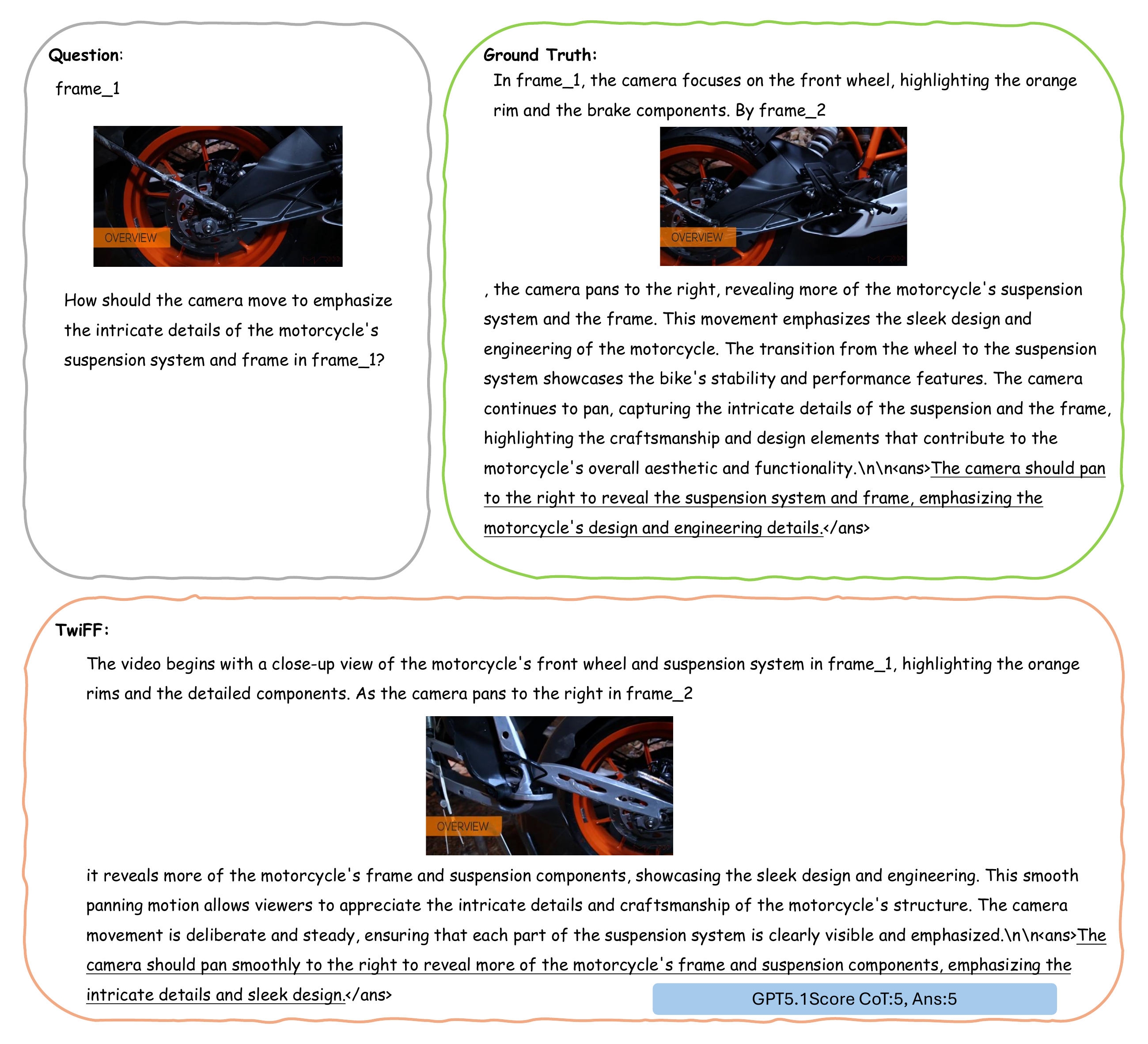}}
  \caption{A sample correct case in TwiFF-Bench camera task.}
  \label{fig:vis_camera_1}
\end{figure}
\begin{figure}[h]
  \centering
  \centerline{\includegraphics[width=0.95\linewidth]{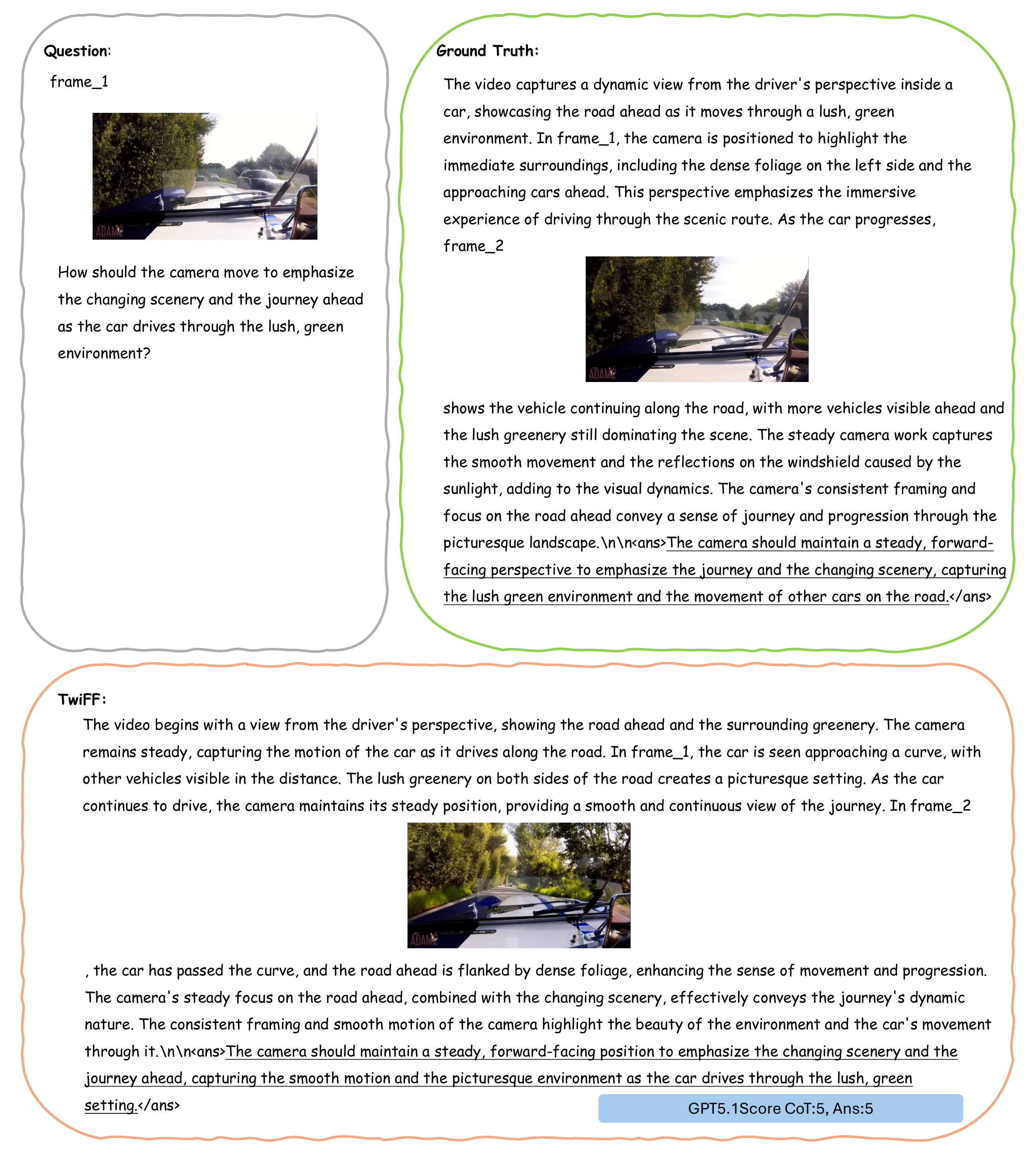}}
  \caption{A sample correct case in TwiFF-Bench camera task.}
  \label{fig:vis_camera_2}
\end{figure}
\begin{figure}[h]
  \centering
  \centerline{\includegraphics[width=0.95\linewidth]{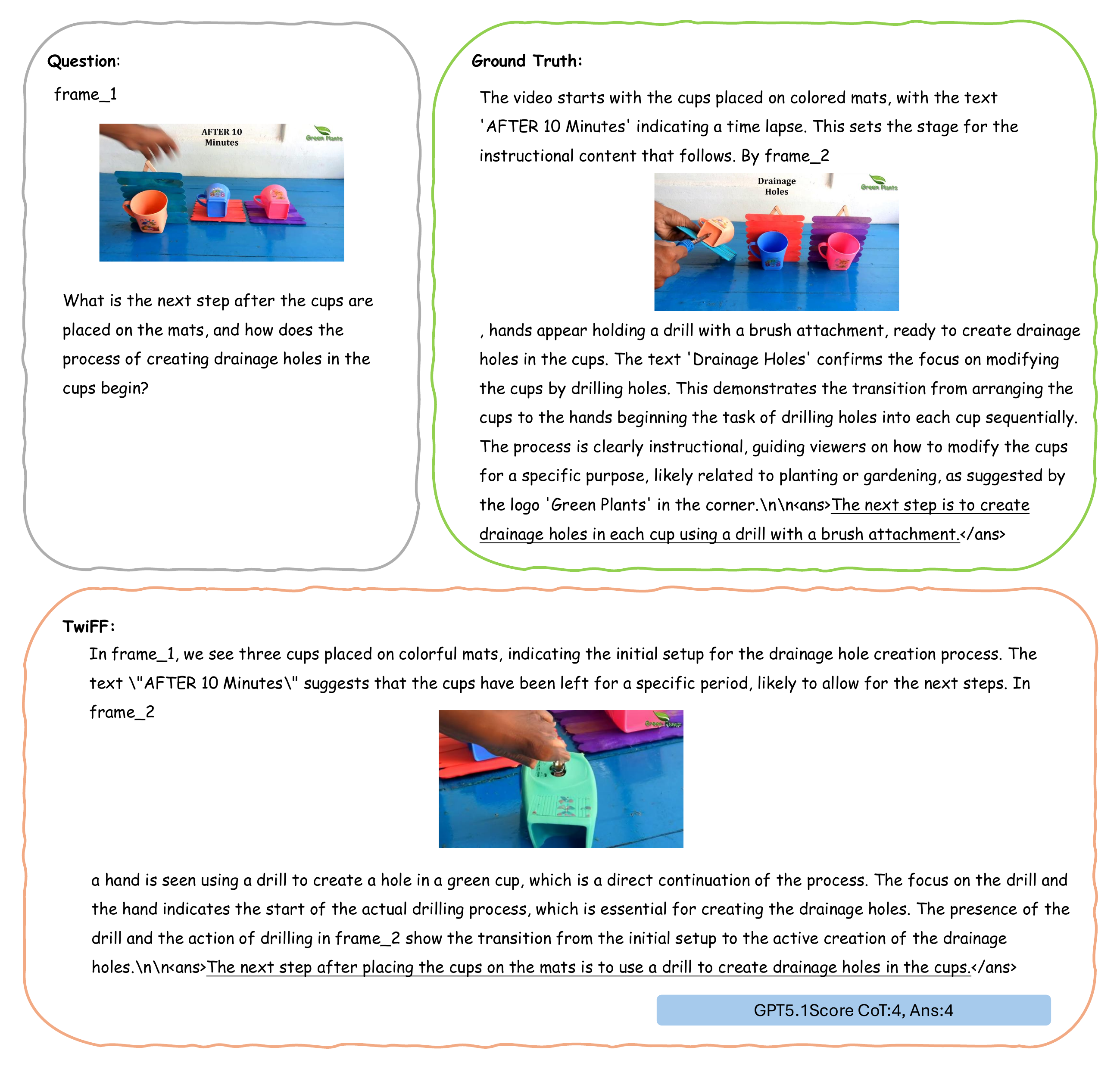}}
  \caption{A sample correct case in TwiFF-Bench instructional task.}
  \label{fig:vis_instructional_1}
\end{figure}
\begin{figure}[h]
  \centering
  \centerline{\includegraphics[width=0.95\linewidth]{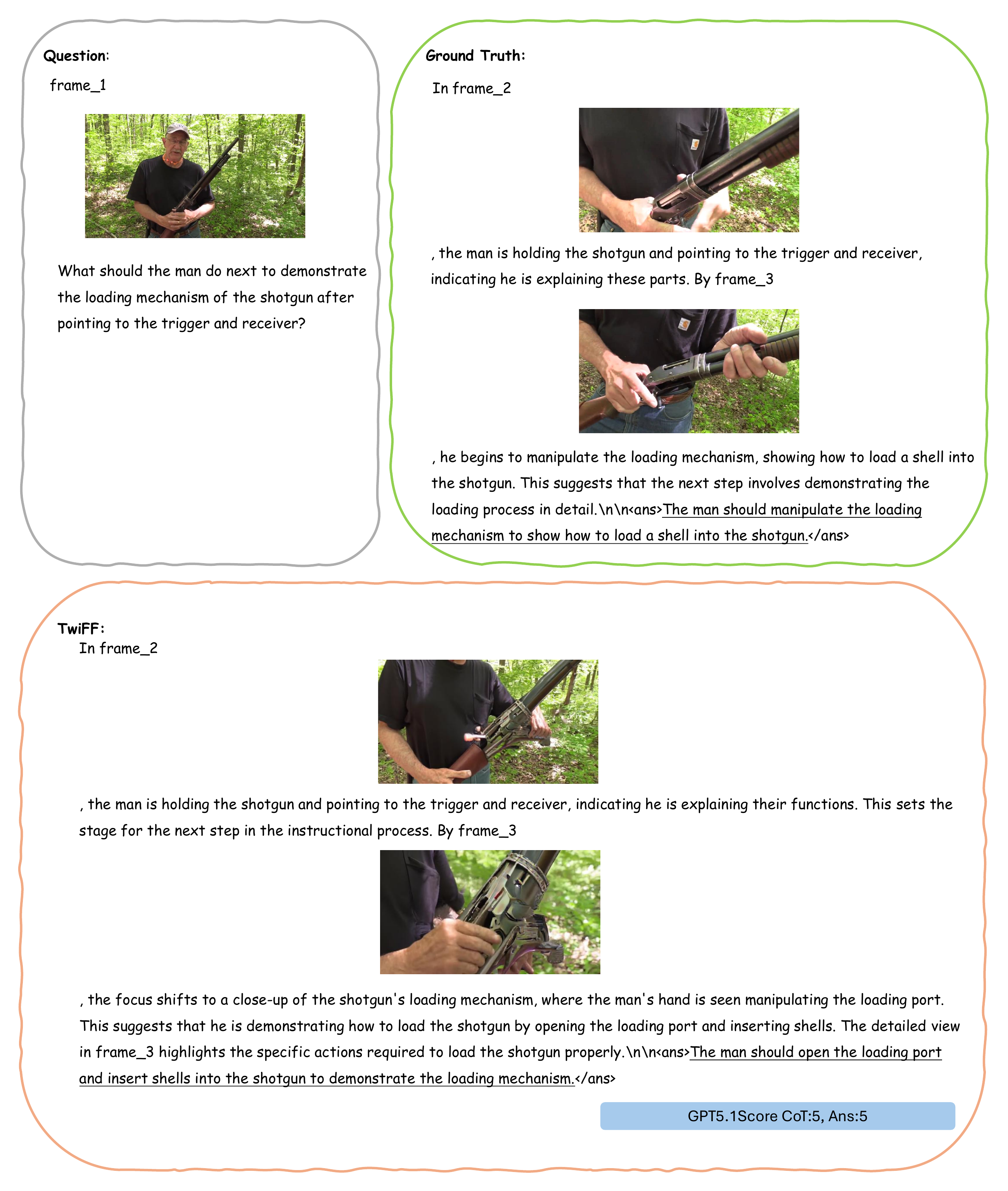}}
  \caption{A sample correct case in TwiFF-Bench instructional task.}
  \label{fig:vis_instructional_2}
\end{figure}
\begin{figure}[h]
  \centering
  \centerline{\includegraphics[width=0.95\linewidth]{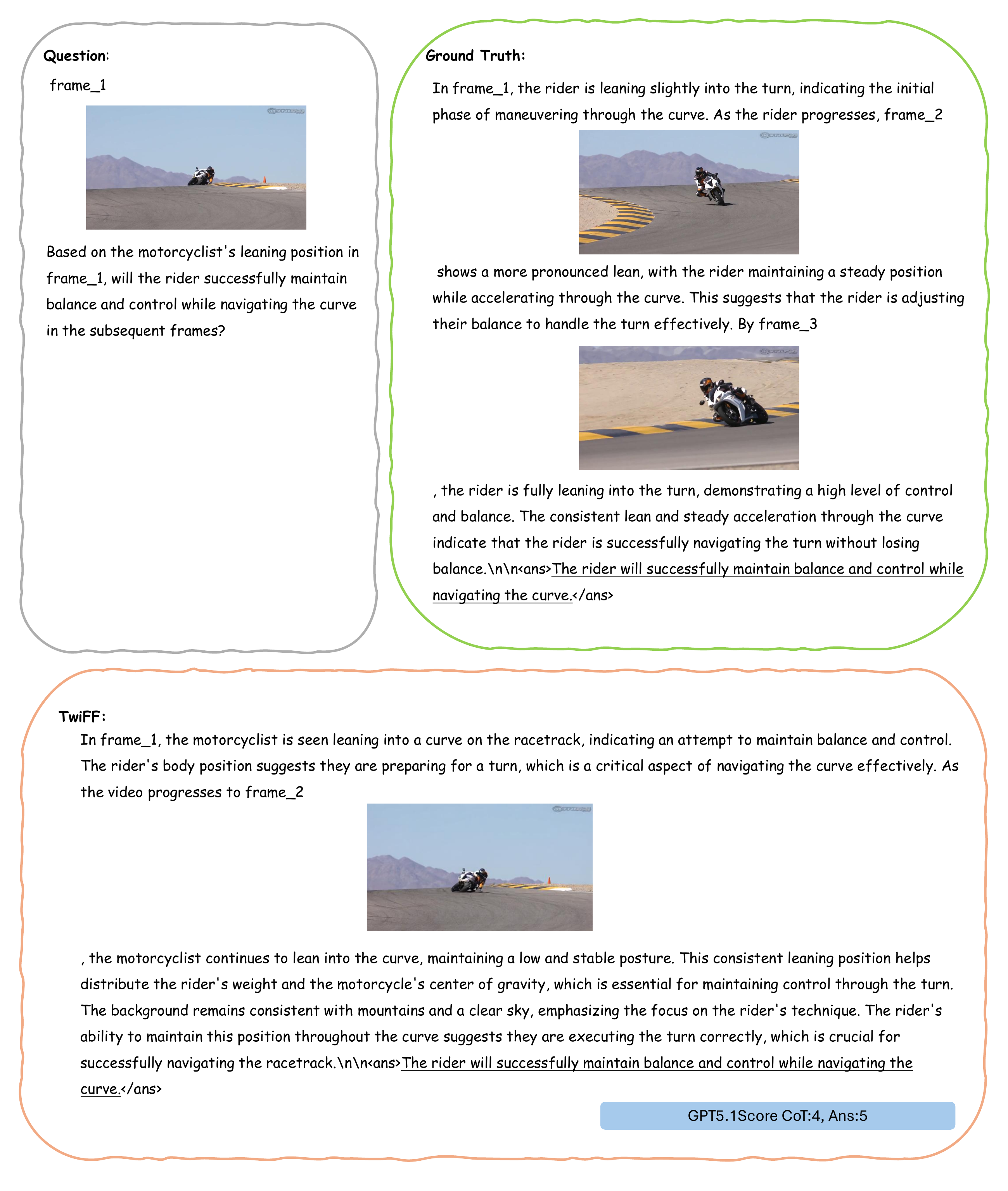}}
  \caption{A sample correct case in TwiFF-Bench predictive task.}
  \label{fig:vis_predictive_1}
\end{figure}
\begin{figure}[h]
  \centering
  \centerline{\includegraphics[width=0.95\linewidth]{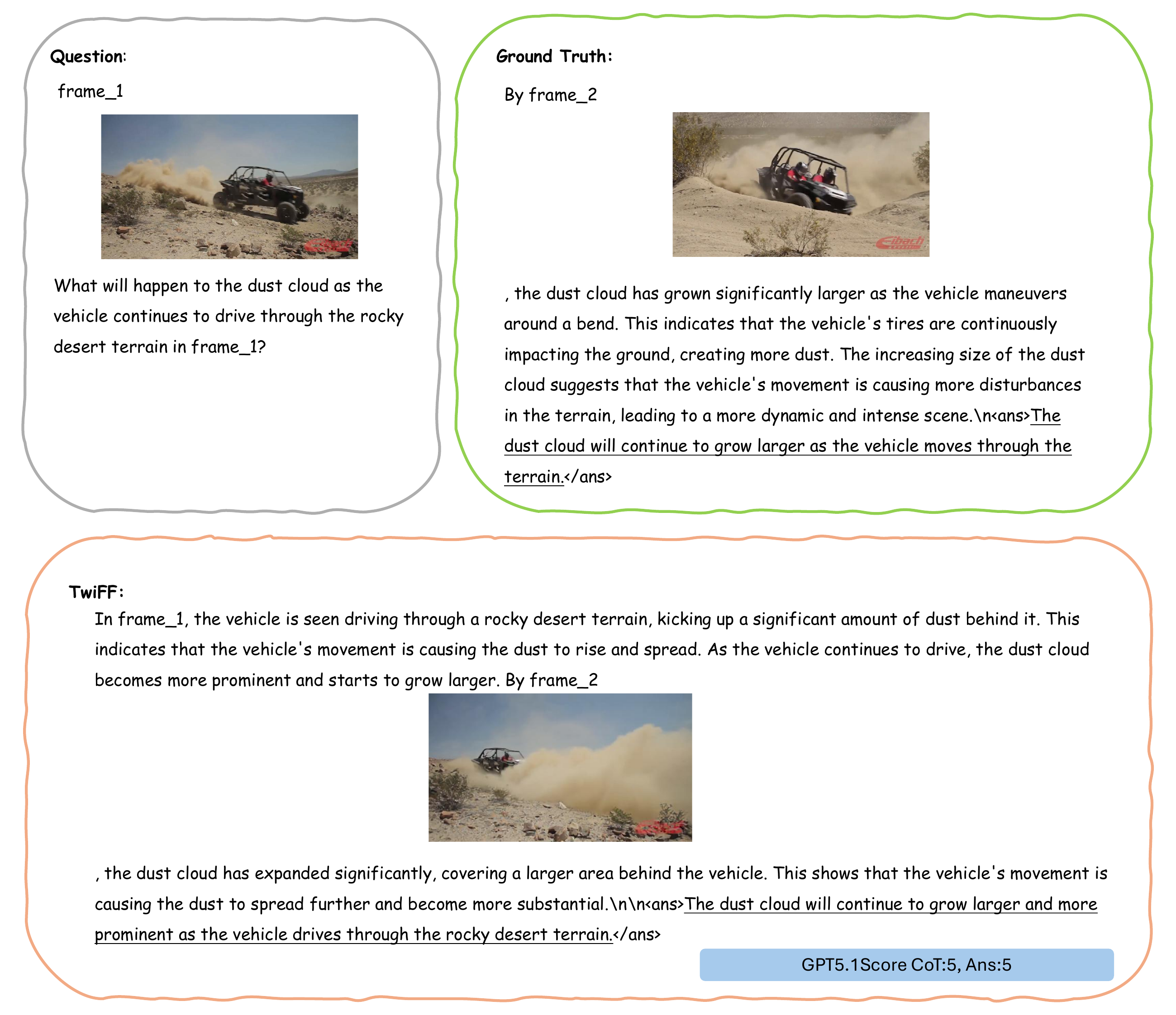}}
  \caption{A sample correct case in TwiFF-Bench predictive task.}
  \label{fig:vis_predictive_2}
\end{figure}
\begin{figure}[h]
  \centering
  \centerline{\includegraphics[width=0.95\linewidth]{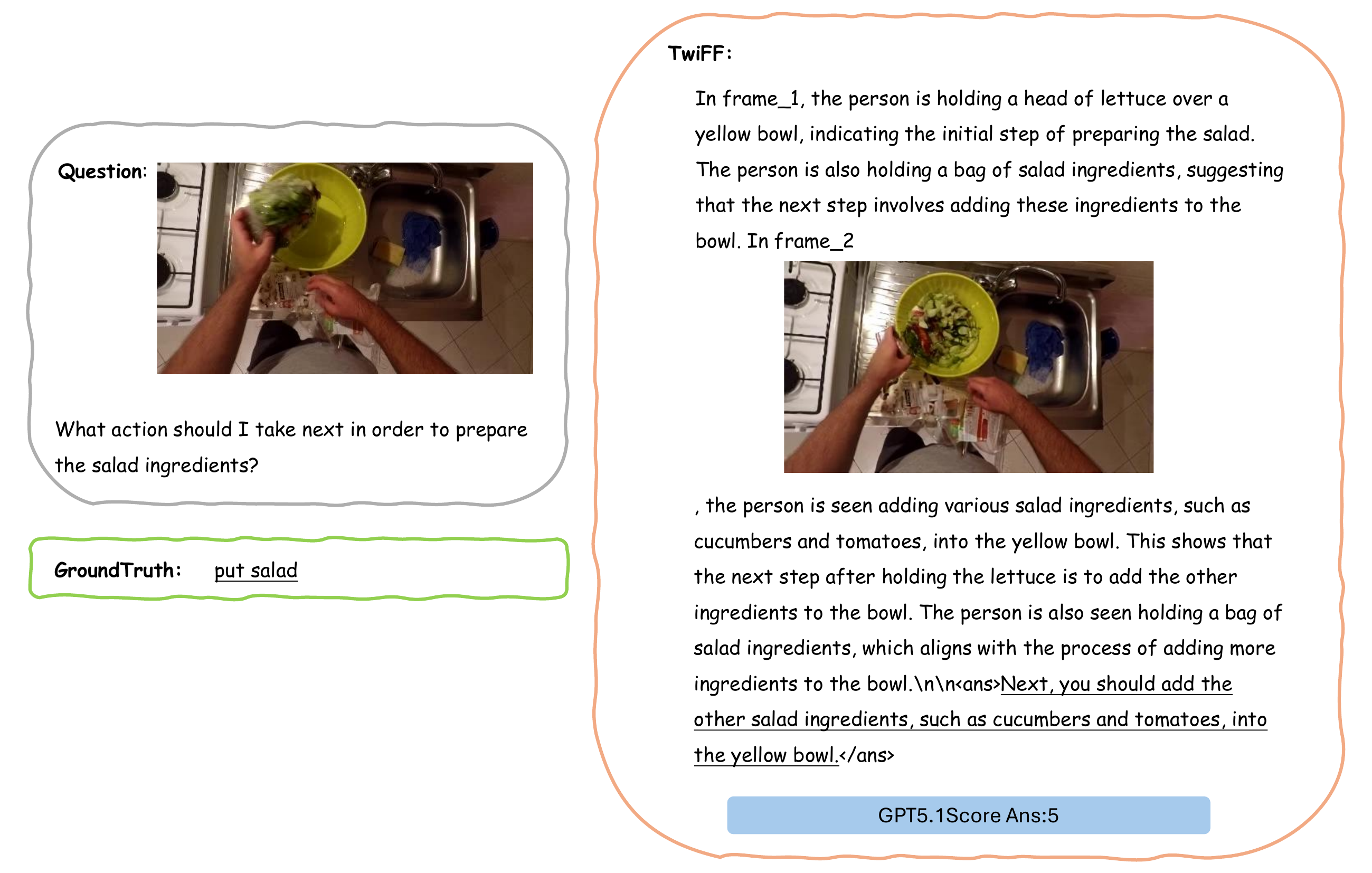}}
  \caption{A sample correct case in Seed-Bench-R1.}
  \label{fig:vis_SBR1_1}
\end{figure}
\begin{figure}[h]
  \centering
  \centerline{\includegraphics[width=0.95\linewidth]{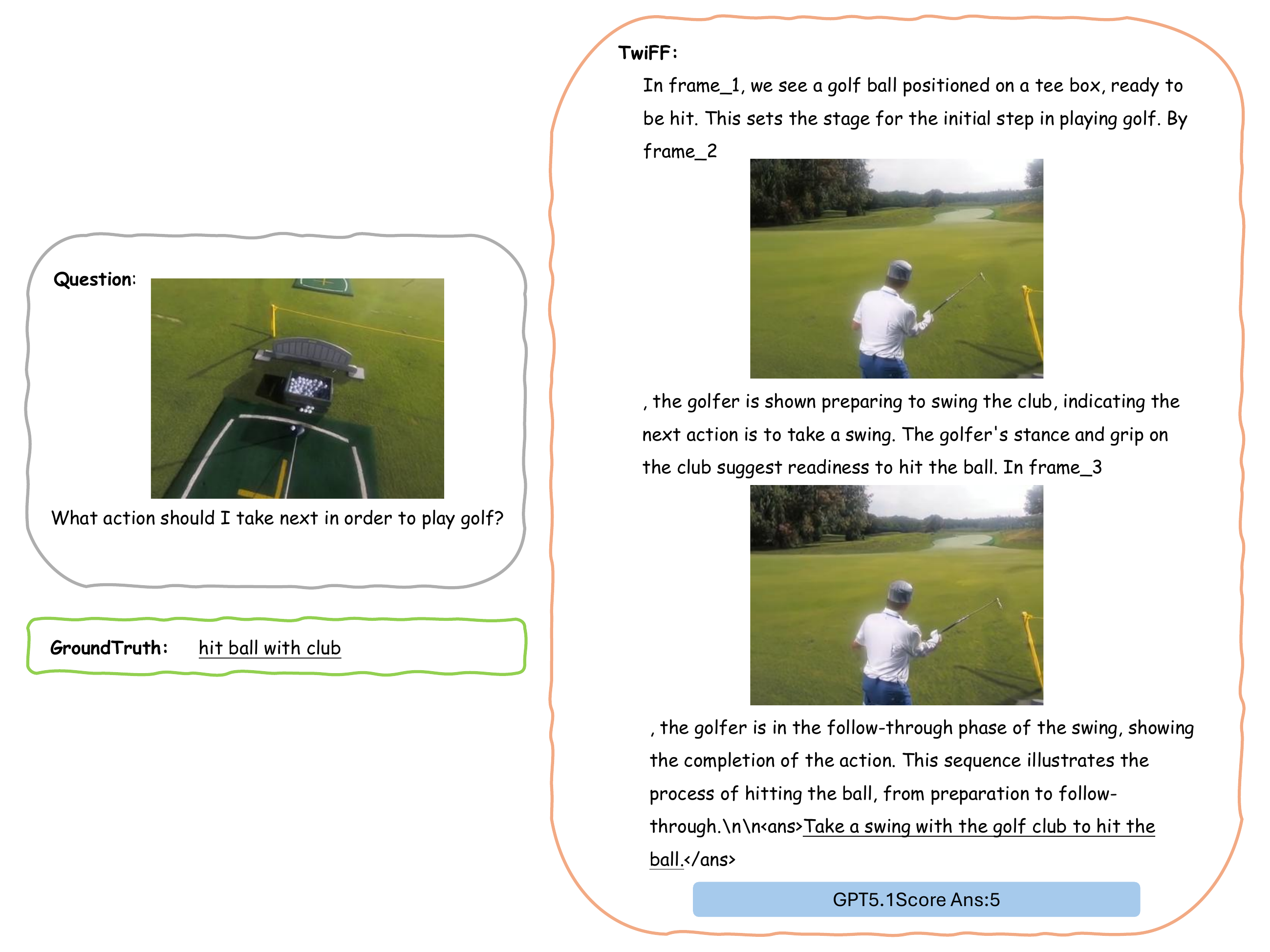}}
  \caption{A sample correct case in Seed-Bench-R1.}
  \label{fig:vis_SBR1_2}
\end{figure}
%%%%%%%%%%%%%%%%%%%%%%%%%%%%%%%%%%%%%%%%%%%%%%%%%%%%%%%%%%%%%%%%%%%%%%%%%%%%%%%
%%%%%%%%%%%%%%%%%%%%%%%%%%%%%%%%%%%%%%%%%%%%%%%%%%%%%%%%%%%%%%%%%%%%%%%%%%%%%%%

\end{document}